\documentclass[journal]{IEEEtran}

\usepackage{cite}

\ifCLASSINFOpdf
  \usepackage[pdftex]{graphicx}
  \graphicspath{{../images/}}
  \DeclareGraphicsExtensions{.pdf,.jpeg,.png}
\else
\fi
\ifCLASSOPTIONcompsoc
 \usepackage[caption=false,font=normalsize,labelfont=sf,textfont=sf]{subfig}
\else
 \usepackage[caption=false,font=footnotesize]{subfig}
\fi

\usepackage{microtype}
\usepackage{amsmath}
\usepackage{booktabs} 
\usepackage{amssymb}
\usepackage{multirow}
\usepackage{enumitem}
\usepackage{amsthm}
\usepackage{algorithm}
\usepackage{algpseudocode}
\usepackage{bm}
\usepackage{adjustbox}
\usepackage{array}
\usepackage{url}
\usepackage{color}
\usepackage{hyperref}
\usepackage{tikz}
\usepackage{lipsum}

\boldmath

\DeclareMathOperator*{\argmin}{arg\,min}

\def\EModel{\xi}
\newcommand{\xbacki}{x_{\backslash i}}	
\newcommand{\scope}[1]{\colorbox{gray!20}{#1}}
\newcommand{\method}[1]{\colorbox{gray!20}{#1}}
\newcommand{\usage}[1]{\colorbox{gray!20}{#1}}

\usetikzlibrary{arrows,shapes,positioning,shadows,trees,calc,decorations.markings}

\tikzset{
  basic/.style   = {draw, text width=2cm, drop shadow, font=\sffamily, rectangle},
  root/.style    = {basic, rounded corners=2pt, thin, align=center, fill=gray!30},
  level 2/.style = {basic, rounded corners=6pt, thin,align=center, fill=gray!50, text width=13em},
  level 3/.style = {basic, thin, align=left, fill=gray!10, text width=13em, node distance=2.1cm and 2cm}
}

\input{timeline}

\begin{document}

\title{Opportunities and Challenges in Explainable Artificial Intelligence (XAI): A Survey}

\author{Arun~Das,~\IEEEmembership{Graduate~Student~Member,~IEEE,}
        and~Paul~Rad,~\IEEEmembership{Senior~Member,~IEEE}
\thanks{This work has been submitted to the IEEE for possible publication. Copyright may be transferred without notice, after which this version may no longer be accessible.}
\thanks{A. Das is with the Department of Electrical and Computer Engineering, University of Texas at San Antonio, San Antonio, TX, 78249 USA. e-mail: arun.das@utsa.edu.}
\thanks{P. Rad is with the Department of Information Systems and Cyber Security, University of Texas at San Antonio, San Antonio, TX, 78249 USA. e-mail: peyman.najafirad@utsa.edu.}%
}

\maketitle

\begin{abstract}
Nowadays, deep neural networks are widely used in mission critical systems such as healthcare, self-driving vehicles, and military which have direct impact on human lives. However, the black-box nature of deep neural networks challenges its use in mission critical applications, raising ethical and judicial concerns inducing lack of trust. Explainable Artificial Intelligence (XAI) is a field of Artificial Intelligence (AI) that promotes a set of tools, techniques, and algorithms that can generate high-quality interpretable, intuitive, human-understandable explanations of AI decisions. In addition to providing a holistic view of the current XAI landscape in deep learning, this paper provides mathematical summaries of seminal work. We start by proposing a taxonomy and categorizing the XAI techniques based on their scope of explanations, methodology behind the algorithms, and explanation level or usage which helps build trustworthy, interpretable, and self-explanatory deep learning models. We then describe the main principles used in XAI research and present the historical timeline for landmark studies in XAI from 2007 to 2020. After explaining each category of algorithms and approaches in detail, we then evaluate the explanation maps generated by eight XAI algorithms on image data, discuss the limitations of this approach, and provide potential future directions to improve XAI evaluation.
\end{abstract}

\begin{IEEEkeywords}
explainable ai, xai, interpretable deep learning, machine learning, computer vision, neural network.
\end{IEEEkeywords}

%
\IEEEpeerreviewmaketitle

\section{Introduction}
\label{sec:introduction}
Artificial Intelligence (AI) based algorithms, especially using deep neural networks, are transforming the way we approach real-world tasks done by humans. Recent years have seen a surge in the use of Machine Learning (ML) algorithms in automating various facets of science, business, and social workflow. The surge is partly due to the uptick of research in a field of ML, called Deep Learning (DL), where thousands (even billions) of neuronal parameters are trained to generalize on carrying out a particular task. Successful use of DL algorithms in healthcare \cite{Torres2018,Lee2019,Chen2020}, ophthalmology \cite{Sayres2019,Das2019,Son2020}, developmental disorders \cite{MohammadianRad2018,Heinsfeld2018,Silva2020Temporal}, in autonomous robots and vehicles \cite{You2019,Grigorescu2019,Feng2020}, in image processing classification and detection \cite{Sahba2018,Bendre2020Human}, in speech and audio processing \cite{Boles2017,Panwar2017}, cyber-security \cite{Parra2020Detecting,Chacon2019Deep}, and many more indicate the reach of DL algorithms in our daily lives. 
Easier access to high-performance compute nodes using cloud computing ecosystems, high-throughput AI accelerators to enhance performance, and access to big-data scale datasets and storage enables deep learning providers to research, test, and operate ML algorithms at scale in small edge devices \cite{Kwasniewska2019}, smartphones \cite{Zhang2019Deep}, and AI-based web-services using Application Programming Interfaces (APIs) for wider exposure to any applications.

The large number of parameters in Deep Neural Networks (DNNs) make them complex to understand and undeniably harder to interpret. Regardless of the cross-validation accuracy or other evaluation parameters which might indicate a good learning performance, deep learning (DL) models could inherently learn or fail to learn representations from the data which a human might consider important. Explaining the decisions made by DNNs require knowledge of the internal operations of DNNs, missing with non-AI-experts and end-users who are more focused on getting accurate solution. Hence, often the ability to interpret AI decisions are deemed secondary in the race to achieve state-of-the-art results or crossing human-level accuracy. 

Recent interest in XAI, even from governments especially with the European General Data Protection Regulation (GDPR) \cite{AIHLEG2019} regulation, shows the important realization of the ethics \cite{Cath2017,Keskinbora2019,Etzioni2017,Bostrom2014,stahl2018ethics}, trust \cite{Weld2019,Lui2018,Hengstler2016}, bias \cite{Chen2019Hidden,Challen2019,Sinz2019,Osoba2017} of AI, as well as the impact of adversarial examples \cite{Kurakin2016,Goodfellow2015,Su2019,Huang2017} in fooling classifier decisions. 
In \cite{Miller2019}, Miller et al. describes that curiosity is one of the primary reason why people ask for explanations to specific decisions. Another reason might be to facilitate better learning - to reiterate model design and generate better results. Each explanation should be consistent across similar data points and generate stable or similar explanation on the same data point over time \cite{Sokol2020}. Explanations should make the AI algorithm expressive to improve human understanding, confidence in decision making, and promote impartial and just decisions. Thus, in order to maintain transparency, trust, and fairness in the ML decision-making process, an explanation or an interpretable solution is required for ML systems.

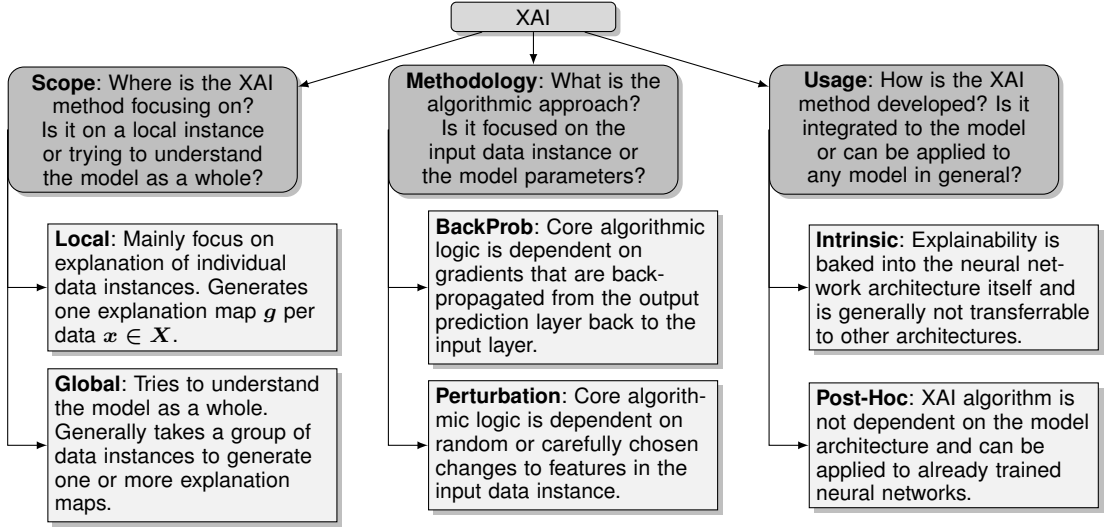
\begin{figure*}[!t]
    \centering
    \footnotesize
    \begin{tikzpicture}[
        level 1/.style={sibling distance=18em},
        edge from parent/.style={->,draw},
        >=latex]
        \node[root] {XAI}
        child {node[level 2] (c1) {\textbf{Scope}: Where is the XAI method focusing on? Is it on a local instance or trying to understand the model as a whole?}}
        child {node[level 2] (c2) {\textbf{Methodology}: What is the algorithmic approach? Is it focused on the input data instance or the model parameters?}}
        child {node[level 2] (c3) {\textbf{Usage}: How is the XAI method developed? Is it integrated to the model or can be applied to any model in general?}};
        \begin{scope}[every node/.style={level 3}]
            \node [below of = c1, xshift=15pt] (c11) {\textbf{Local}: Mainly focus on explanation of individual data instances. Generates one explanation map \bm{$g$} per data \bm{$x \in X$}.};
            \node [below of = c11] (c12) {\textbf{Global}: Tries to understand the model as a whole. Generally takes a group of data instances to generate one or more explanation maps.};
            \node [below of = c2, xshift=15pt] (c21) {\textbf{BackProb}: Core algorithmic logic is dependent on gradients that are backpropagated from the output prediction layer back to the input layer.};
            \node [below of = c21] (c22) {\textbf{Perturbation}: Core algorithmic logic is dependent on random or carefully chosen changes to features in the input data instance.};
            \node [below of = c3, xshift=15pt] (c31) {\textbf{Intrinsic}: Explainability is baked into the neural network architecture itself and is generally not transferrable to other architectures.};
            \node [below of = c31] (c32) {\textbf{Post-Hoc}: XAI algorithm is not dependent on the model architecture and can be applied to already trained neural networks.};
        \end{scope}
        \foreach \value in {1,2}
          \draw[->] (c1.west) |- (c1\value.west);
        \foreach \value in {1,2}
          \draw[->] (c2.west) |- (c2\value.west);
        \foreach \value in {1,2}
          \draw[->] (c3.west) |- (c3\value.west);
    \end{tikzpicture}
    \caption{General categorization of the survey in terms of scope, methodology, and usage.}
    \label{fig:surveycategory}
\end{figure*}

An explanation is a way to verify the output decision made by an AI agent or algorithm. For a cancer detection model using microscopic images, an explanation might mean a map of input pixels which contribute to the model output. For a speech recognition model, an explanation might be the power spectrum information during a specific time which contributed more towards the current output decision. Explanations can be also based on the parameters or activations of the trained models explained either by using surrogates such as decision trees or by using gradients or other methods. In the context of reinforcement learning algorithms, an explanation might be given as to why an agent made a certain decision over another. However, the definitions of interpretable and explainable AI are often generic and might be misleading \cite{Rudin2019} and should integrate some form of reasoning \cite{Doran2018}. 

A collection of AI models, such as decision-trees and rule-based models, is inherently interpretable. However, there are affected by the drawbacks of Interpretability-versus-Accuracy trade-off compared to the Deep Learning models.  This paper discusses the different approaches and perspectives of researchers to address the problem of the explainability of deep learning algorithms. Methods can be used effectively if the model parameters and architecture are already known. However, modern API-based AI services produce more challenges because of the relative `black-box' \cite{Castelvecchi2016} nature of the problem where the end-user has information only on the input provided to the deep learning model and not the model itself. 

In this survey, we present a comprehensive overview of explainable and interpretable algorithms with a timeline of important events and research publications into three well-defined taxonomies as illustrated in Figure \ref{fig:surveycategory}. Unlike many other surveys which only categorize and summarize the published research in a high-level, we provide additional mathematical overviews and algorithms of seminal works in the field of XAI. The algorithms presented in the survey are clustered into three well-defined categories which are described in detail in the following sections. Various evaluation techniques for XAI presented in literature are also discussed along with discussion on the limitations and future directions of these methods. 

Our contributions can be summarized as the following:

\begin{enumerate}
    \item In order to systematically analyze explainable and interpretable algorithms in deep learning, we taxonomize XAI to three well-defined categories to improve clarity and accessibility of the approaches.
    \item We examine, summarize and classify the core mathematical model and algorithms of recent XAI research on the proposed taxonomy and discuss the timeline for seminal work.
    \item We generate and compare the explanation maps for eight different XAI algorithms, outline the limitations of this approach, and discuss potential future directions to improve trust, transparency, and bias and fairness using deep neural network explanations.
\end{enumerate}

Our survey is based on published research, from the year 2007 to 2020, from various search sources including Google Scholar, ACM Digital Library, IEEEXplore, ScienceDirect, Spinger, and preprints from arXiv. Keywords such as \textit{explainable artificial intelligence}, \textit{XAI}, \textit{explainable machine learning}, \textit{explainable deep learning}, \textit{interpretable machine learning} were used as search parameters.

\section{Taxonomies and Organization}
Prior published survey's on general explainability have classified XAI techniques based on scope and usage \cite{Adadi2018}. Key differences of this survey are classification based on methodology behind the XAI algorithms for deep learning, focus on mathematical summaries of the seminal papers, and evaluation strategies for XAI algorithms. We also mention popular open-source software implementations of various algorithms described in this survey. We summarize the taxonomies discussed in the survey in this section based on the illustration provided in Figure \ref{fig:surveycategory}:

\begin{itemize}
    \item \textbf{Scope}: Scope of explanations can be either \textit{local} or \textit{global}. Some methods can be extended to both. Locally explainable methods are designed to express, in general, the individual feature attributions of a single instance of input data \bm{$x$} from the data population \bm{$X$}. For example, given a text document and a model to understand the sentiment of text, a locally explainable model might generate attribution scores for individual words in the text. Globally explainable models provide insight into the decision of the model as a whole - leading to an understanding about attributions for an array of input data. Local and global scope of explanations are described in detail in Section \ref{sec:scope}.

    \item \textbf{Methodology}: Core algorithmic concept behind the explainable model can generally be categorized based on the methodology of implementation. In general, both local and global explainable algorithms can be categorized as either \textit{backpropagation-based} or \textit{perturbation-based} methods. In backpropagation-based methods, the explainable algorithm does one or more forward pass through the neural network and generates attributions during the backpropagation stage utilizing partial derivatives of the activations. Examples include saliency maps, saliency relevance maps, and class activation maps.

    Perturbation-based explainable algorithms focus on perturbing the feature set of a given input instance by either using occlusion, partially substituting features using filling operations or generative algorithms, masking, conditional sampling, etc. Here, generally, only forward pass is enough to generate the attribution representations without the need for backpropagating gradients. These methodology differences are described in Section \ref{sec:methodology}.
   
    \item \textbf{Usage}: A well developed explainable method with a specific scope and methodology can be either embedded to the neural network model itself or applied as an external algorithm for explanation. Any explainable algorithm which is dependent on the model architecture fall into the \textit{model-intrinsic} category. Most model-intrinsic algorithms are model-specific such that any change in the architecture will need significant changes in the method itself or minor changes of hyperparameters of the explainable algorithm. 

    Generally, significant research interest is seen in developing model-agnostic \textit{post-hoc} explanations, where the predictions of an already existing well-performing neural network model can be explained using ad-hoc explainable methods. Post-hoc methods are also widely applied in variety of input modalities such as images, text, tabular data, etc. These differences in the 'usage' of explainability methods are described in Section \ref{sec:usage}.
\end{itemize}

In Section \ref{sec:evaluation}, we discuss some of the evaluation strategies used to qualitatively or quantitatively evaluate the performance of XAI algorithms discussed in this survey. We present a list of desirable constraints applicable to XAI algorithms to improve its real-world performance as well as expressiveness in terms of transparency, trust, and bias understanding. These desirable qualities can be used as a guide to generate novel XAI algorithms which is favorable as well as expressive. Our study suggests that the evaluation methods are still immature and have an enormous potential for further research. We also provide a list of popular software packages that are open-sourced in GitHub platform. We chose the packages with considerable user support and implemented algorithms. All software platforms supports explaining either Scikit-Learn, Tensorflow, or PyTorch machine learning models. After describing the evaluation methods and software packages, we conclude our survey in Section \ref{sec:conclusion}.

In our survey, all mathematical equations and algorithms described are based on a set of notations as described in Table \ref{tab:notation}. The mathematical equations described in the survey might be different from their respective research publications as we have used similar notations to describe the same mathematical idea throughout the survey. This is done to aid the readers and have a common repository of notations. Also, a timeline of seminal research in the field is illustrated in Figure \ref{fig:timeline}. The timeline provides information such as the name of the XAI method, name of first author, and year of publication.

\begin{table}[!t]
\caption{Table of notations}
\label{tab:notation}
\centering

\begin{tabular}{||l||p{0.6\columnwidth}||}
\hline
 Notation & Description \\
 \hline
 \hline
\bm{$x$                     } & Single instance of input data from a population \bm{$X$} \\
\bm{$\xbacki$               } & Set of all input features except \bm{$i$}th feature \\
\bm{$y$                     } & Class label of input \bm{$x$} from a population \bm{$Y$} \\
\bm{$\bar{y}$               } & Predicted label for input \bm{$x$} \\
\bm{$x_i$                   } & \bm{$i$}th feature in input instance \bm{$x$}\\
\bm{$x^{(i)}$               } & Single instance from \bm{$X$} at location \bm{$i$} \\
\bm{$y^{(i)}$               } & Single instance from \bm{$Y$} at location \bm{$i$} \\
\bm{$\bar{y}^{(i)}$         } & Predicted label for input \bm{$x^{(i)}$} \\
\bm{$f(.)$                  } & Neural network model \\
\bm{$\theta$                } & Parameters of the neural network \\
\bm{$G$                     } & Explanation function \\                
\bm{$g$                     } & Explanation of a model \bm{$f$} \\
\bm{$z_{i,j}(\theta, x)$    } & Activation output of node or layer \bm{$j$} for feature \bm{$i$} \\
\bm{$z_j$                   } & Activation output summary of node or layer \bm{$j$} for all features \\
\bm{$x^{*}$                 } & Activation map of input \bm{$x$} \\
\bm{$x^{'}$                 } & Binary activation map of input \bm{$x$} \\
\bm{$S_c$                   } & Class score function \\
\bm{$Z^{'}$                 } & Coalition vector for SHAP \\
\bm{$M$                     } & Maximum coalition size \\
\bm{$\phi_j \in \mathbb{R}$ } & Feature attribution for feature \bm{$j$} \\
\bm{$R(z_j)$                } & Relevance of activation \bm{$z_j$} \\
\bm{$I$                     } & Input image \\
\hline
\hline

\end{tabular}
\end{table}

\section{Definitions and Preliminaries}
\label{sec:definitions}
\begin{figure}[ht]
\begin{center}
\includegraphics[width=\linewidth]{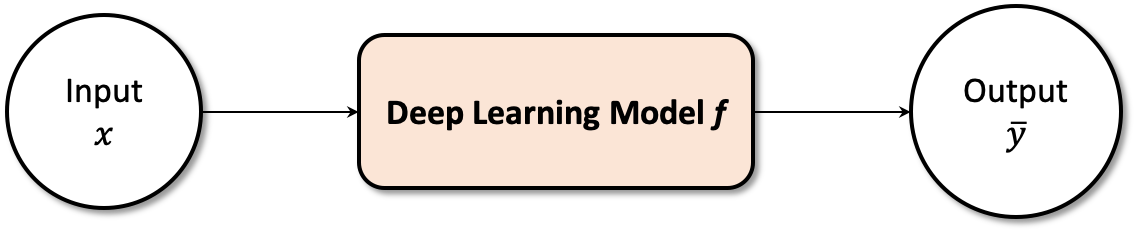}
\end{center}
  \caption{High-level illustration of deep learning model \bm{$f$}. Generally, a single input instance \bm{$x$} generates outputs \bm{$\Bar{y}$}. No other metadata or explanations are generated other than the output classification. Most model inference scenarios involve this method where model \bm{$f$} is considered as a blob of information which takes an input \bm{$x$} and generates an output \bm{$\Bar{y}$}}
\label{fig:generalmodel}
\end{figure}

Various prior publications debate the nuances in defining Explainability and Interpretability of neural networks \cite{Dosilovic2018,Chakraborty2017}. We support the general concept of explainable AI as a suite of techniques and algorithms designed to improve the trustworthiness and transparency of AI systems. Explanations are described as extra metadata information from the AI model that offers insight into a specific AI decision or the internal functionality of the AI model as a whole. 
Various Explainability approaches applied to Deep Neural Networks are presented in this literature survey. Figure \ref{fig:generalmodel} illustrates one such deep learning model which takes one input and generates one output prediction. Goal of explainable algorithms applied to deep neural networks are towards explaining these predictions using various methods summarized in this survey.

Generally, for an input \bm{$x \in {\mathbb R}^d$}, a deep learning model function \bm{$f(\theta)$} describes \bm{$f: {\mathbb R}^d \rightarrow {\mathbb R}^C$}, where \bm{$C$} is the number of output classes and \bm{$\theta$} the parameters of the model in a classification problem. Now, the model inference can be described as \bm{$\Bar{y} = f(\theta, x)$} where \bm{$\Bar{y}$} is the output prediction. 
We now define the key concepts explored in the survey, namely explainability of deep learning models. Further sections of the survey explain these definitions in finer detail.

\textbf{Definition 1: } \textit{\textbf{Interpretability} is a desirable quality or feature of an algorithm which provides enough expressive data to understand how the algorithm works.} 

Here, interpretable domains could include images or text which are comprehensible by humans. Cambridge Dictionary defines: ``\textit{If something is interpretable, it is possible to find its meaning or possible to find a particular meaning in it}". 

\textbf{Definition 2: } \textit{\textbf{Interpretation} is a simplified representation of a complex domain, such as outputs generated by a machine learning model, to meaningful concepts which are human-understandable and reasonable.}

Output predictions of a simple rule-based model can be easily interpreted by traversing the rule-set. Similarly a small decision tree can be easily understood. Or the Deep Convolution Networks (CNN) model that can identify the parts of the input image that led to the decision.

\textbf{Definition 3: } \textit{An \textbf{explanation} is additional meta information, generated by an external algorithm or by the machine learning model itself, to describe the feature importance or relevance of an input instance towards a particular output classification.}

For a deep learning model \bm{$f$} with input \bm{$x$} and output prediction of \bm{$\Bar{y}$} of class \bm{$c$}, an explanation \bm{$g$} can be generated, generally as an explanation map \bm{$E$}, where \bm{$E: {\mathbb R}^d \rightarrow {\mathbb R}^d$}. Here, \bm{$g$} is an object of same shape as the input which describes the feature importance or relevance of that particular dimension to the class output. For an image, the explanation map can be an equally sized pixel map whereas for text, it might be word-by-word influence scores. 

\textbf{Definition 4: } \textit{For a deep learning model \bm{$f$}, if the model parameters \bm{$\theta$} and the model architecture information are known, the model is considered a \textbf{white-box}.}

A white-box model improves model-debugging and promotes trust. However, knowing the model architecture and parameters alone won't make the model explainable.

\textbf{Definition 5: } \textit{A deep learning model \bm{$f$} is considered a \textbf{black-box} if the model parameters and network architectures are hidden from the end-user.}

Typically, deep learning models served on web-based services or restricted business platforms are exposed using APIs which takes an input form the user and provides the model result as text, visual, or auditory presentation respective to the expected model output \bm{$\Bar{y}$}.

\subsection{Why Is Research on XAI Important?}
With the use of AI algorithms in healthcare \cite{Zhu2019}, credit scoring \cite{VanThiel2019}, loan acceptance \cite{Turiel2019}, and more, the need to explain an ML model result is important for ethical, judicial, as well as safety reasons. Even though there are different facets to why XAI is important, our study suggests that the most important concerns are three-fold: 1) trustability, 2) transparency, and 3) bias and fairness of AI algorithms. Current business models include interpretation as a step before serving the ML models on production systems, however are often limited to small tree-based models. With the use of highly non-linear deep learning algorithms with millions of parameters in ML pipelines, XAI techniques must improve all three concerns mentioned above.

\begin{figure}[!t]
\begin{center}
\includegraphics[width=0.95\linewidth]{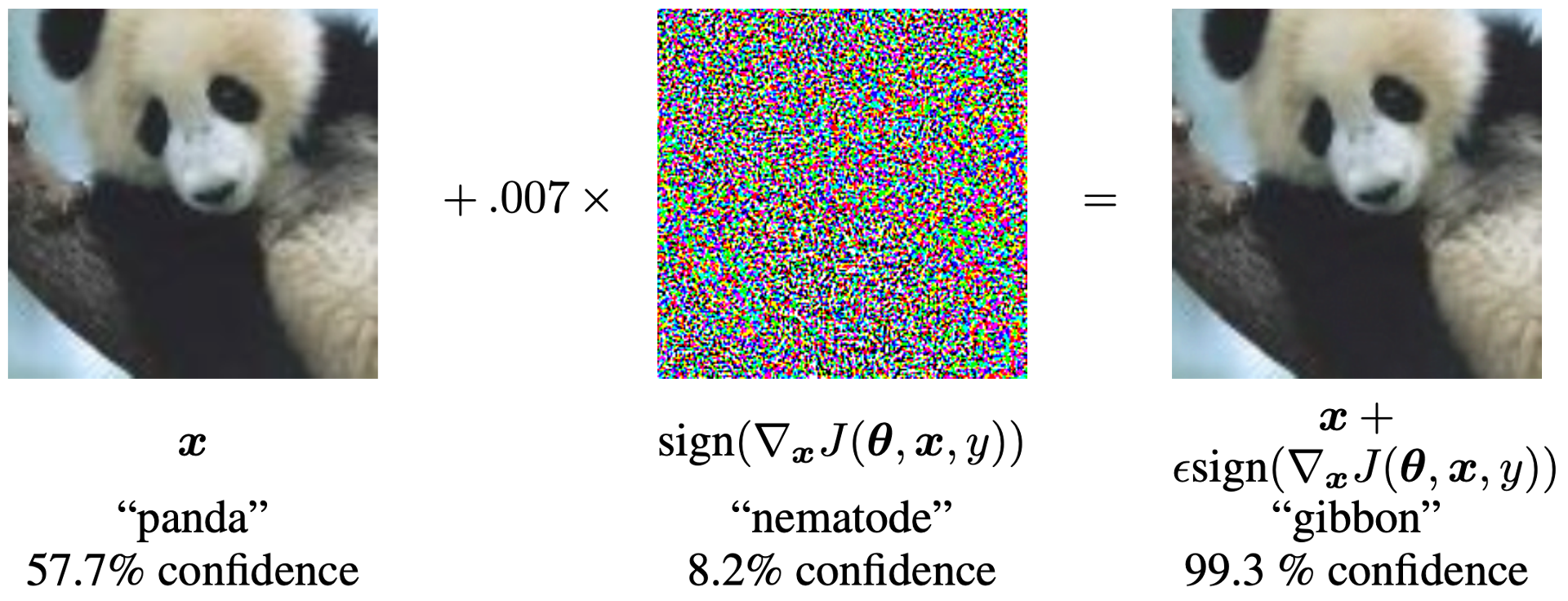}
\end{center}
\caption{Illustration from \cite{Goodfellow2015} showing an adversarial attack where an image class Panda is deliberately attacked to predict as a Gibbon with high confidence. Note that the attacked image is visually similar to the original image and humans are unable to understand any changes.}
\label{fig:adversarial}
\end{figure}

\begin{figure}[!t]
\begin{center}
\includegraphics[width=0.95\linewidth]{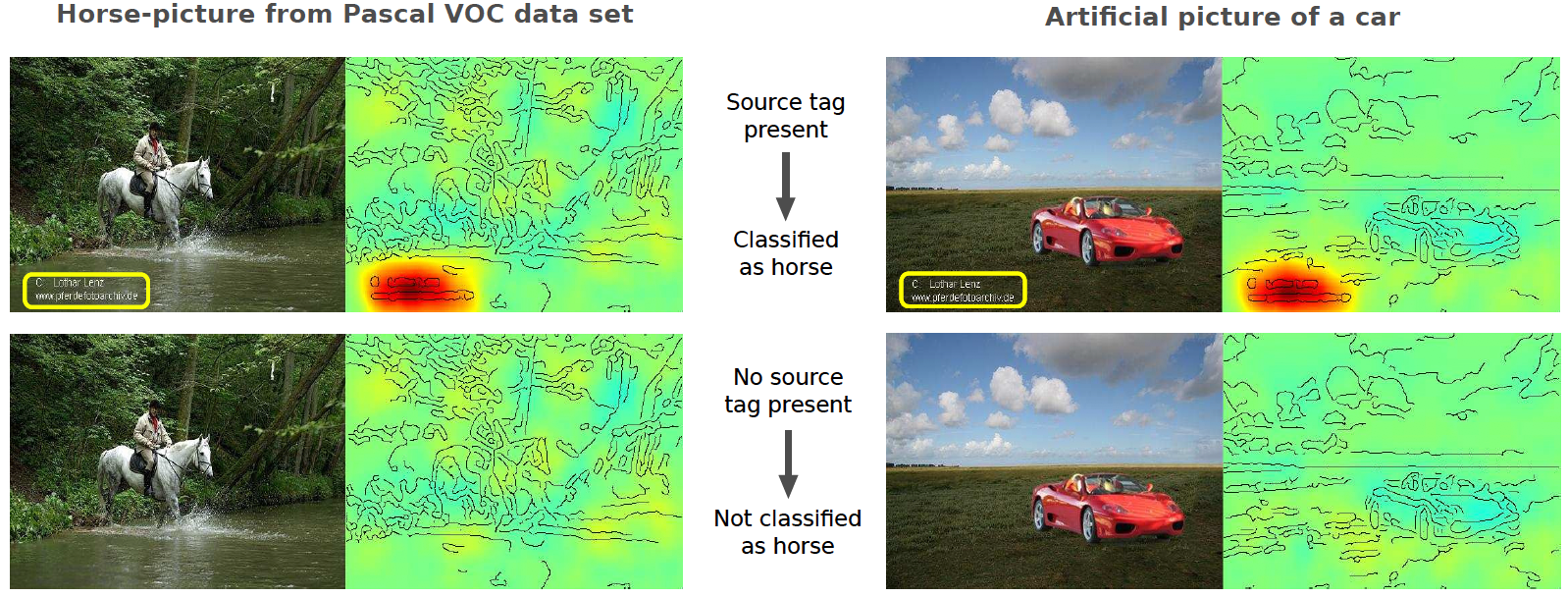}
\end{center}
\caption{Illustration from \cite{Lapuschkin2019} showing how text in images can fool classifiers into believing that the text is a feature for a particular task.}
\label{fig:sourcetagmiss}
\end{figure}

\begin{enumerate}
    \item \textbf{Improves Transparency}: XAI improves transparency and fairness by creating a human-understandable justification to the decisions and could find and deter adversarial examples \cite{Goodfellow2015} if used properly. 
    
    \textbf{Definition 6: } \textit{A deep learning model is considered \textbf{transparent} if it is expressive enough to be human-understandable. Here, transparency can be a part of the algorithm itself or using external means such as model decomposition or simulations.}
    
    Transparency is important to assess the quality of output predictions and to ward off adversaries. An adversarial example could hinder accurate decision making capabilities of a classifier by fooling the classifier into believing that a fake image is infact real. Figure \ref{fig:adversarial} illustrates such an example where an image of a Panda is predicted as a Gibbon with high confidence after the original Panda image was tampered by adding some adversarial noise. Figure \ref{fig:sourcetagmiss} illustrates a classifier learning to classify based on text data such as source tags or watermarks in advertisements in images. As we rely more on autonomous algorithms to aid our daily lives, quality of AI algorithms to mitigate attacks \cite{Kurakin2016} and provide transparency in terms of model understanding, textual, or visual reports should be of prime importance. 
    
    \item \textbf{Improves Trust}: As a social animal, our social lives, decisions, and judgements are primarily based on the knowledge and available explanations to situations and the trust we generate. A scientific explanation or logical reasoning for a sub-optimal decision is better than a highly confident decision without any explanations.

\begin{figure}[!t]
\begin{center}
\includegraphics[width=0.99\linewidth]{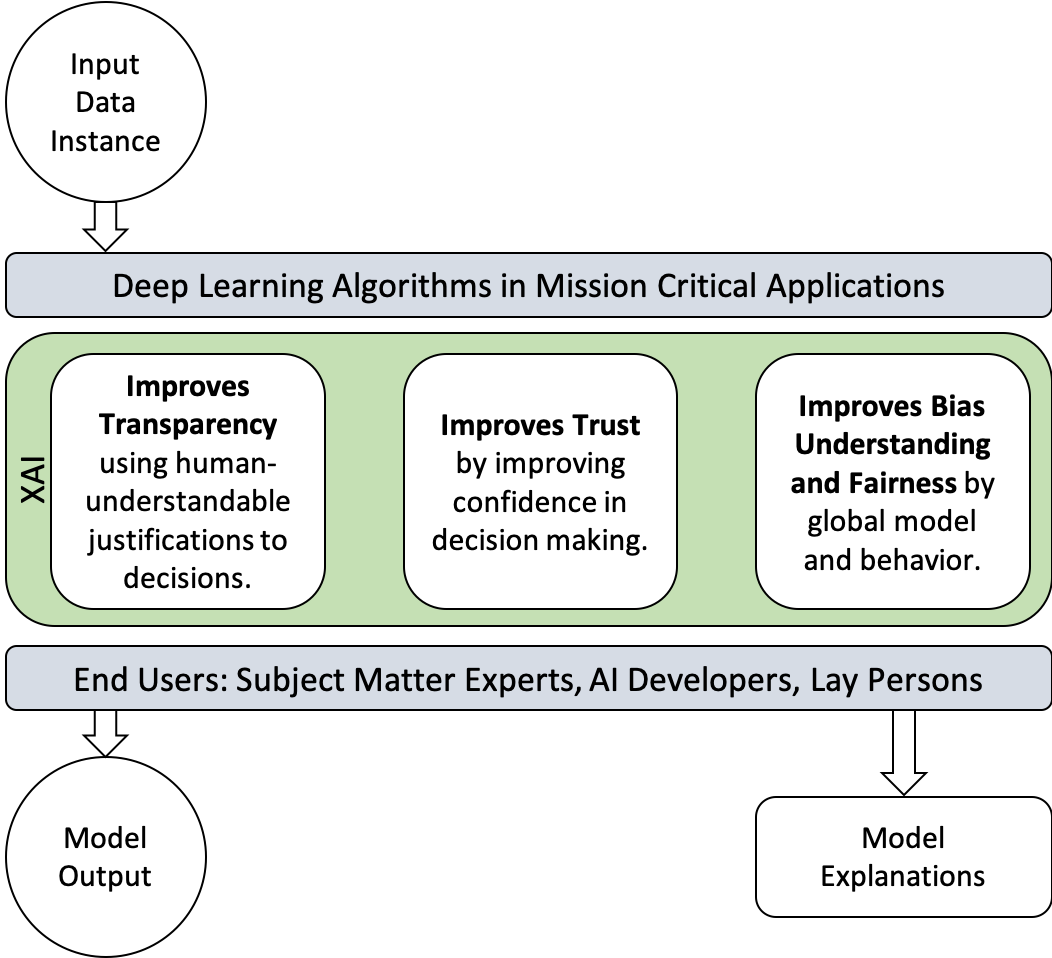}
\end{center}
\caption{Significant expected improvements when using XAI techniques to support decision making of end-users. We believe XAI is important due to improvements in trust, transparency, and in understanding bias and fairness.}
\label{fig:whyxai}
\end{figure}

\begin{figure}[!t]
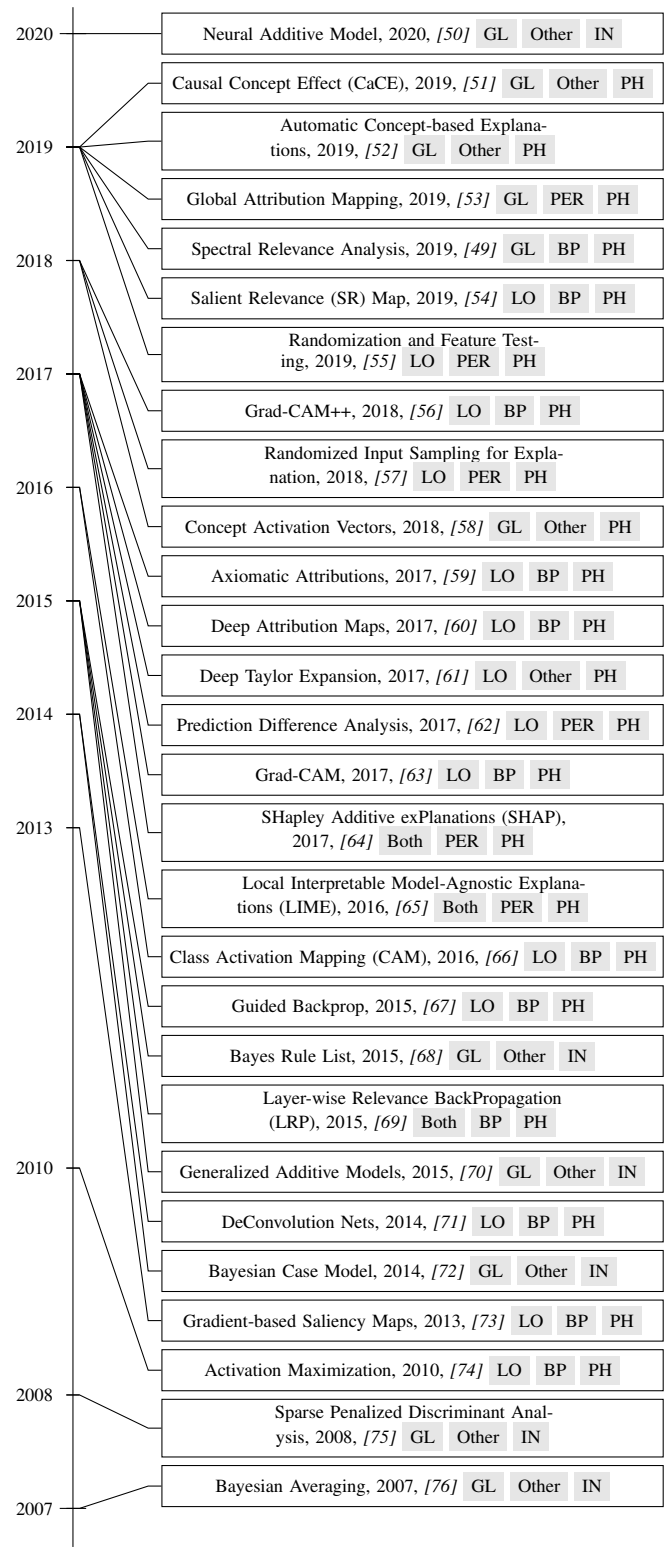

\scriptsize{
\begin{timeline}{2020}{2007}{0.5cm}{1cm}{6.5cm}{20.5cm}
\entry{2020}{Neural Additive Model, 2020,                       \textit{\cite{Agarwal2020Neural}}       \scope{GL}   \method{Other}  \usage{IN}}
\entry{2019}{Causal Concept Effect (CaCE), 2019,                \textit{\cite{Goyal2019Explaining}}     \scope{GL}   \method{Other}  \usage{PH}}
\entry{2019}{Automatic Concept-based Explanations, 2019,        \textit{\cite{Ghorbani2019Towards}}     \scope{GL}   \method{Other}  \usage{PH}}
\entry{2019}{Global Attribution Mapping, 2019,                  \textit{\cite{Ibrahim2019Global}}       \scope{GL}   \method{PER}    \usage{PH}}
\entry{2019}{Spectral Relevance Analysis, 2019,         \textit{\cite{Lapuschkin2019}}          \scope{GL}   \method{BP}     \usage{PH}}
\entry{2019}{Salient Relevance (SR) Map, 2019,                  \textit{\cite{Li2019}}                  \scope{LO}   \method{BP}     \usage{PH}}
\entry{2019}{Randomization and Feature Testing, 2019,           \textit{\cite{Burns2019}}               \scope{LO}   \method{PER}    \usage{PH}}
\entry{2018}{Grad-CAM++, 2018,                                  \textit{\cite{Chattopadhay2018}}        \scope{LO}   \method{BP}     \usage{PH}}
\entry{2018}{Randomized Input Sampling for Explanation, 2018,   \textit{\cite{Petsiuk2018}}             \scope{LO}   \method{PER}    \usage{PH}}
\entry{2018}{Concept Activation Vectors, 2018,                  \textit{\cite{Kim2018Interpretability}} \scope{GL}   \method{Other}  \usage{PH}}
\entry{2017}{Axiomatic Attributions, 2017,                      \textit{\cite{Sundararajan2017}}        \scope{LO}   \method{BP}     \usage{PH}}
\entry{2017}{Deep Attribution Maps, 2017,                       \textit{\cite{Ancona2017}}              \scope{LO}   \method{BP}     \usage{PH}}
\entry{2017}{Deep Taylor Expansion, 2017,                       \textit{\cite{Montavon2017}}            \scope{LO}   \method{Other}  \usage{PH}}
\entry{2017}{Prediction Difference Analysis, 2017,              \textit{\cite{Zintgraf2017}}            \scope{LO}   \method{PER}    \usage{PH}}
\entry{2017}{Grad-CAM, 2017,                                    \textit{\cite{Selvaraju2017}}           \scope{LO}   \method{BP}     \usage{PH}}
\entry{2017}{SHapley Additive exPlanations (SHAP), 2017,        \textit{\cite{Lundberg2017}}            \scope{Both} \method{PER}    \usage{PH}}
\entry{2016}{Local Interpretable Model-Agnostic Explanations (LIME), 2016,    \textit{\cite{Ribeiro2016}}             \scope{Both} \method{PER}    \usage{PH}}
\entry{2016}{Class Activation Mapping (CAM), 2016,              \textit{\cite{Zhou2016Learning}}        \scope{LO}   \method{BP}     \usage{PH}}
\entry{2015}{Guided Backprop, 2015,                             \textit{\cite{Springenberg2015}}        \scope{LO}   \method{BP}     \usage{PH}}
\entry{2015}{Bayes Rule List, 2015,                             \textit{\cite{Letham2015}}              \scope{GL}   \method{Other}  \usage{IN}}
\entry{2015}{Layer-wise Relevance BackPropagation (LRP), 2015,  \textit{\cite{Bach2015}}                \scope{Both} \method{BP}     \usage{PH}}
\entry{2015}{Generalized Additive Models, 2015,           \textit{\cite{Caruana2015}}             \scope{GL}   \method{Other}  \usage{IN}}
\entry{2014}{DeConvolution Nets, 2014,                          \textit{\cite{Zeiler2014}}              \scope{LO}   \method{BP}     \usage{PH}}
\entry{2014}{Bayesian Case Model, 2014,                         \textit{\cite{Kim2014}}                 \scope{GL}   \method{Other}  \usage{IN}}
\entry{2013}{Gradient-based Saliency Maps, 2013,                \textit{\cite{Simonyan2013}}            \scope{LO}   \method{BP}     \usage{PH}}
\entry{2010}{Activation Maximization, 2010,                     \textit{\cite{Erhan2010}}               \scope{LO}   \method{BP}     \usage{PH}}
\entry{2008}{Sparse Penalized Discriminant Analysis, 2008,      \textit{\cite{Grosenick2008}}           \scope{GL}   \method{Other}  \usage{IN}}
\entry{2007}{Bayesian Averaging, 2007,                          \textit{\cite{Schetinin2007}}           \scope{GL}   \method{Other}  \usage{IN}}
\end{timeline}
}
    \caption{A timeline of seminal works towards explainable AI algorithms is illustrated. The grey highlights indicate \textbf{scope} (GL: global, LO: local, Both: GL and LO), \textbf{methodology} (BP: backprop, PER: perturbation, Other: neither BP or PER, and \textbf{usage} level (IN: intrinsic or PH: post-hoc) of the algorithms. }
    \label{fig:timeline}
\end{figure}

\begin{table}[!t]
\caption{Table of Abbreviations}
\label{tab:abbrv}
\renewcommand{\arraystretch}{1.15}
\centering

\begin{tabular}{||l|l||}
\hline
Abbreviation & Definition                                      \\ \hline \hline
ACE          & Automatic Concept-based Explanations            \\ \hline
AI           & Artificial Intelligence                         \\ \hline
API          & Application Programming Interface               \\ \hline
BAM          & Benchmarking Attribution Methods                \\ \hline
BRL          & Bayesian Rule List                              \\ \hline
CaCE         & Causal Concept Effect                           \\ \hline
CAM          & Class Activation Mapping                        \\ \hline
CAV          & Concept Activation Vectors                      \\ \hline
CNN          & Convolutional Neural Network                    \\ \hline
DeConvNet    & Deconvolution Neural Network                    \\ \hline
DL           & Deep Learning                                   \\ \hline
DNN          & Deep Neural Network                             \\ \hline
EG           & Expected Gradients                              \\ \hline
FMRI         & Functional Magnetic Resonance Imaging           \\ \hline
GAM          & Generalized Additive Models                     \\ \hline
IG           & Integrated Gradients                            \\ \hline
IRT          & Interpretability Randomization Test             \\ \hline
LIME         & Local Interpretable Model-Agnostic Explanations \\ \hline
LRP          & Layer-wise Relevance BackPropagation            \\ \hline
ML           & Machine Learning                                \\ \hline
NAM          & Neural Additive Models                          \\ \hline
OSFT         & One-Shot Feature Test                           \\ \hline
ReLU         & Rectified Linear Unit                           \\ \hline
RISE         & Randomized Input Sampling for Explanation       \\ \hline
RNN          & Recurrent Neural Network                        \\ \hline
SCS          & System Causability Scale                        \\ \hline
SHAP         & SHapley Additive exPlanations                   \\ \hline
SPDA         & Sparse Penalized Discriminant Analysis          \\ \hline
SpRAy        & Spectral Relevance Analysis                     \\ \hline
SR           & Salient Relevance                               \\ \hline
TCAV         & Testing with Concept Activation Vectors         \\ \hline
t-SNE        & t-Stochastic Neighbor Embedding                 \\ \hline
VAE          & Variational Auto Encoders                       \\ \hline
XAI          & Explainable Artificial Intelligence             \\ \hline
\end{tabular}
\end{table}

    \textbf{Definition 7: } \textit{\textbf{Trustability} of deep learning models is a measure of confidence, as humans, as end-users, in the intended working of a given model in dynamic real-world environments.}
    
     Thus, `Why a particular decision was made' is of prime importance to improve the trust \cite{Rossi2019} of end-users including subject matter experts, developers, law-makers, and laypersons alike \cite{King2020,Cerka2015,Arrieta2020Explainable}. Fundamental explanations to classifier prediction is ever so important to stake-holders and government agencies to build trustability as we transition to a connected AI-driven socio-economic environment.
    
    \item \textbf{Improves Model Bias Understanding and Fairness}: XAI promotes fairness and helps mitigate biases introduced to the AI decision either from input datasets or poor neural network architecture. 
    
    \textbf{Definition 8: } \textit{\textbf{Bias} in deep learning algorithms indicate the disproportionate weight, prejudice, favor, or inclination of the learnt model towards subsets of data due to both inherent biases in human data collection and deficiencies in the learning algorithm.}
    
    Learning the model behavior using XAI techniques for different input data distributions could improve our understanding of the skewness and biases in the input data. This could generate a robust AI model \cite{Zou2018}. Understanding the input space could help us invest in bias mitigation methods and promote fairer models. 
    
    \textbf{Definition 9: } \textit{\textbf{Fairness} in deep learning is the quality of a learnt model in providing impartial and just decisions without favoring any populations in the input data distribution.}
    
    XAI techniques could be used as a way to improve the expressiveness and generate meaningful explanations to feature correlations for many subspaces in the data distribution to understand fairness in AI. By tracing back the output prediction discriminations back to the input using XAI techniques, we can understand the subset of features correlated to particular class-wise decisions \cite{Du2019Fairness}.
    
\end{enumerate}

As we discussed previously, the use of XAI could provide a software-engineering design on AI with a continuously evolving model based on prior parameters, explanations, issues, and improvements to overall design thereby reducing human bias. However, choosing the right methods for explanation should be done with care, while considering to bake-in interpretability to machine learning models \cite{Rudin2019}. We now proceed with detailed discussions as per the taxonomies.

\section{Scope of Explanation}
\label{sec:scope}
\subsection{Local Explanations}
\begin{figure}[!t]
\begin{center}
\includegraphics[width=0.95\linewidth]{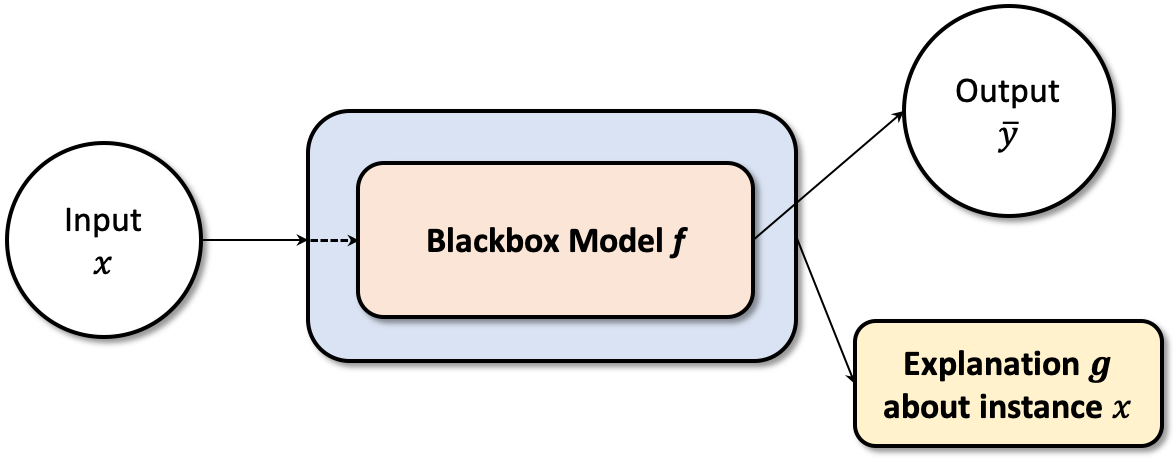}
\end{center}
  \caption{High-level illustration of locally explainable models. Generally, a single input instance is used for explanations.}
\label{fig:localmodel}
\end{figure}

Consider a scenario where a doctor has to make a decision based on the results of a classifier output. The doctor needs careful understanding of the model predictions and concrete answers to the `Why this decision?' question which requires an explanation of the local data point under scrutiny. This level of explaining individual decisions made by a classifier is categorized under locally explainable algorithms. Generally, locally explainable methods focus on a single input data instance to generate explanations by utilizing the different data features.  Here, we are interested in generating \bm{$g$} for explaining the decisions made by \bm{$f$} for a single input instance \bm{$x$}. A high-level diagram is illustrated in Figure \ref{fig:localmodel}.

Founding research in local explanations used heatmaps, rule-based methods, Bayesian techniques, and feature importance matrices to understand feature correlations and importance towards output predictions. The output explanations were always positive real-valued matrices or vectors. Newer research in local explainable models improves the old methods by attribution maps, graph-based, and game-theory based models in which we get a feature-wise score of positive and negative correlations towards an output classification. Here, a positive attribution value means that the particular feature improves output class probability and a negative value means the feature decreased the output class probability. Mathematical equations described in this section and the rest of the survey follows notations tabulated in Table \ref{tab:notation}.

\subsubsection{Activation Maximization}\label{subsec:actmax}
Interpreting a layer-wise feature importance of a Convolutional Neural Network (CNN) model is simpler in the first layer which generally learns the high-level textures and edges. However, as we move deeper into the CNN, importance of specific layers towards a particular prediction is hard to summarize and visualize since parameters of subsequent layers are influenced by that of the previous layers. Hence, preliminary research tried to understand the neuronal activations to input instances as well as individual filters of specific layers.

In 2010, a locally explainable method called Activation Maximization was introduced by Erhan et al. \cite{Erhan2010}, with focus on input patterns which maximize a given hidden unit activation. Here, the authors set an optimization problem of maximizing the activation of a unit. If \bm{$\theta$} is the parameters of the model, \bm{$z_{i,j}(\theta,x)$} is the activation of a particular unit \bm{$i$} from layer \bm{$j$}. By assuming fixed parameters \bm{$\theta$}, an activation map can be found as:

\begin{equation}
x^{*}=\underset{x~s.t.~\|x\|=\rho}{\arg \max} z_{i j}(\theta, x)
\end{equation}

After the optimization converges, we could either find an average of all local minima's to find an explanation map \bm{$g$} or pick the one which maximizes the activations. Here, the goal is to minimize the activation maximization loss by finding larger filter activations correlated to specific input patterns. Thus, we could understand a layer-wise feature importance to an input instance. 
It was one of the first published research to express feature importance of deep learning models and was later improved by many researchers.

\subsubsection{Saliency Map Visualization}\label{subsec:saliencymap}
\begin{figure}[!t]
\begin{center}
\includegraphics[width=0.8\linewidth]{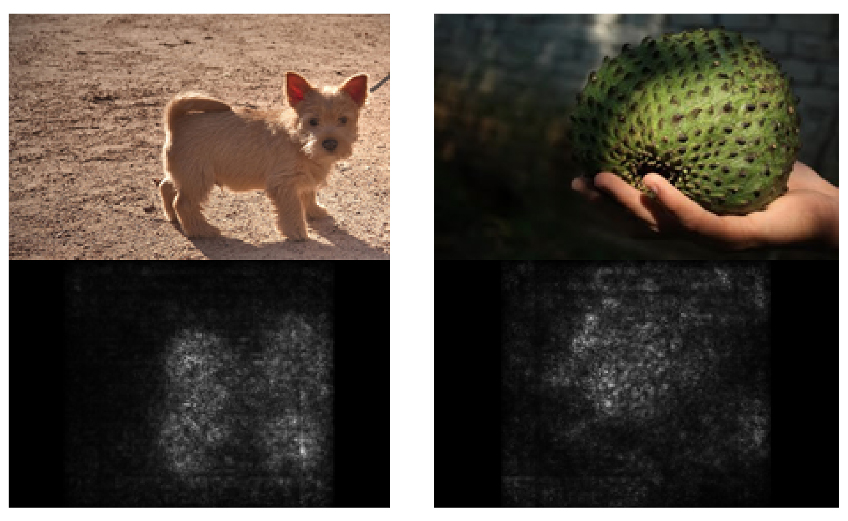}
\end{center}
  \caption{Image-specific class saliency maps using gradient based attribution method is shown. Image courtesy \cite{Simonyan2013}.}
\label{fig:simnonyan_grad}
\end{figure}

Saliency map generation in deep neural networks were first introduced by Simonyan et al. \cite{Simonyan2013} as a way of computing the gradient of the output class category with respect to an input image. By visualizing the gradients, a fair summary of pixel importance can be achieved by studying the positive gradients which had more influence to the output. Authors introduced two techniques of visualization: 1) class model visualizations and 2) image-specific class visualizations as illustrated in Figure \ref{fig:simnonyan_grad}. 

We discuss class model visualization under the global explainable methods in Section \ref{subsec:global}. Image-specific class saliency visualization technique tries to find an approximate class score function \bm{$S_c(I)$}, where \bm{$x$} is the input image with a label class \bm{$c$} using first-order Taylor expansion: 
\begin{equation}
S_{c}(I) \approx w^{T} x+b
\end{equation}
where $w$ is the derivative of the class score function \bm{$S_c$} with respect to the input image \bm{$x$} at a specific point in the image $x_{0}$ such that:
\begin{equation}
w=\left.\frac{\partial S_{c}}{\partial x}\right|_{x_{0}}
\end{equation}

Here, with light image processing, we can visualize the saliency map with respect to the location of input pixels with positive gradients.

\subsubsection{Layer-wise Relevance BackPropagation (LRP)}\label{subsec:LRP}
LRP technique introduced in 2015 by Bach et al. \cite{Bach2015} is used to find relevance scores for individual features in the input data by decomposing the output predictions of the DNN.
The relevance score for each atomic input is calculated by backpropagating the class scores of an output class node towards the input layer. The propagation follows a strict conservation property whereby a equal redistribution of relevance received by a neuron must be enforced. In CNNs, LRP backpropagates information regarding relevance of output class back to input layer, layer-by-layer. In Recurrent Neural Networks (RNNs), relevance is propagated to hidden states and memory cell. Zero relevance is assigned to gates of the RNN. If we consider a simple neural network with input instance \bm{$x$}, a linear output \bm{$y$}, and activation output \bm{$z$}, the system can be described as:

\begin{equation}
\begin{array}{l}
y_{j}=\sum_{i} w_{i j} x_i +b_{j} \\
z_{j}=f\left(y_{j}\right)
\end{array}
\end{equation}

If we consider \bm{$R(z_j)$} as the relevance of activation output, the goal is to get \bm{$R_{i\leftarrow j}$}, that is to distribute \bm{$R(z_j)$} to the corresponding input \bm{$x$}:

\begin{equation}
R_{i \leftarrow j}=R\left(z_{j}\right) \frac{x_i w_{i j}}{y_{j}+\epsilon \operatorname{sign}\left(y_{j}\right)}
\end{equation}

Final relevance score of individual input \bm{$x$} is the summation of all relevance from \bm{$z_j$} for input \bm{$x_i$}:

\begin{equation}
R\left(x\right)=\sum_{j} R_{i \leftarrow j}
\end{equation}

The LRP method have been recently extended to learn the global explainability by using LRP explanation maps as an input to global attribution algorithms. We discuss some such models in section \ref{subsec:global}. Newer research has also shown the importance of using methods such as LRP for model specific operations such as network pruning \cite{Yeom2019Pruning}. Here, authors prune the least important weights or filters of a model by understanding the feature attributions of individual layer. This reduces the computation and storage cost of the AI models without significant drop in the model accuracy. This shows another aspect of using AI in understanding the model behavior and utilizing the new knowledge to improve model performance.

\subsubsection{Local Interpretable Model-Agnostic Explanations (LIME)}
\label{subsec:limelocal}
In 2016, Ribeiro et al. introduced Local Interpretable Model-Agnostic Explanations (LIME) \cite{Ribeiro2016}. To derive a representation that is understandable by humans, LIME tries to find importance of contiguous superpixels (a patch of pixels) in a source image towards the output class. Hence, LIME finds a binary vector \bm{$x^{'} \in \{0,1\}$} to represent the presence or absence of a continuous path or 'superpixel' that provides the highest representation towards class output. This works on a patch-level on a single data input. Hence, the method falls under local explanations. There is also a global explanation model based on LIME called SP-LIME described in the global explainable model sub section. Here, we focus on local explanations.

\begin{algorithm}[!b]
    \small
    \caption{LIME algorithm for local explanations}
    \label{alg:limelocal}
\begin{algorithmic}[1]
    \Statex \textbf{Input:} classifier \bm{$f$}, input sample \bm{$x$}, number of superpixels \bm{$n$}, number of features to pick \bm{$m$}
    \Statex \textbf{Output:} explainable coefficients from the linear model
    
    \State \bm{$\Bar{y} \leftarrow f.\texttt{predict(}x\texttt{)}$}
    
    \For {i in \bm{$n$}}
        \State \bm{$p_i \leftarrow \texttt{Permute(}x\texttt{)}$} \Comment{Randomly pick superpixels}
        \State \bm{$obs_i \leftarrow f.\texttt{predict(}p\texttt{)}$}
        \State \bm{$dist_i \leftarrow |\Bar{y}-obs_i|$}
    \EndFor
    
    \State \bm{$simscore \leftarrow  \texttt{SimilarityScore(}dist\texttt{)}$}
    \State \bm{$x_{pick} \leftarrow \texttt{Pick(}p, simscore, m\texttt{)}$}
    \State \bm{$L \leftarrow \texttt{LinearModel.fit(}p,m,simscore\texttt{)}$}
    \State return \bm{$L$}.weights
    
\end{algorithmic}
\end{algorithm}

\begin{figure*}[!t]
\begin{center}
\includegraphics[width=0.8\linewidth]{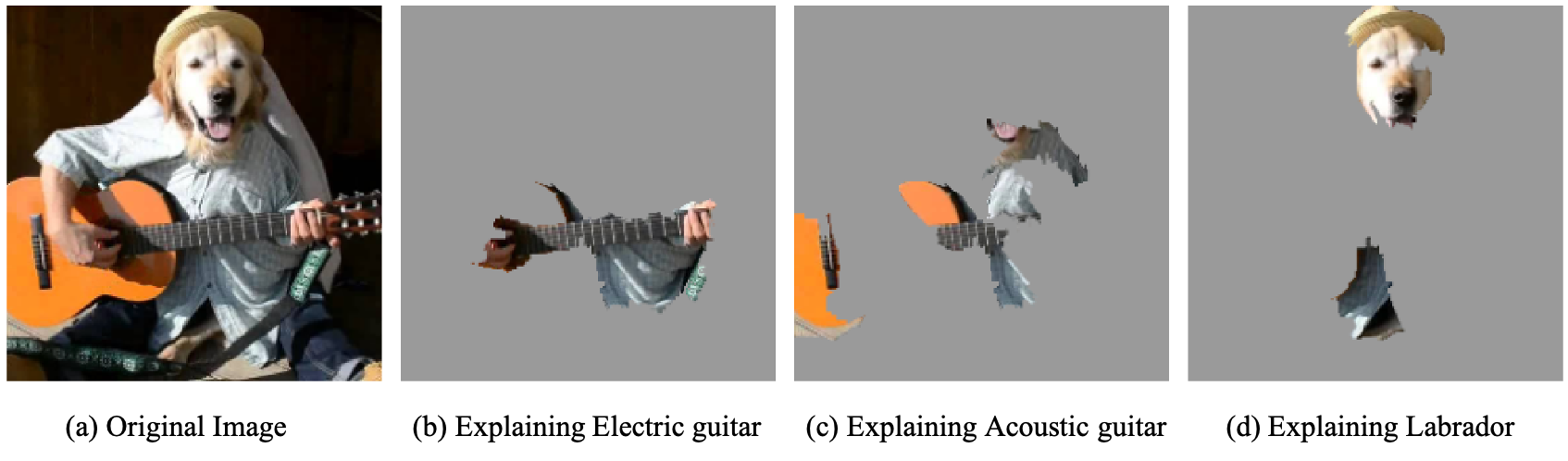}
\end{center}
  \caption{Local explanations of an image classification prediction described using LIME \cite{Ribeiro2016}. Here, top three classes are "electric guitar" $(p=0.32)$, "acoustic guitar" $(p=0.24)$ and "labrador" $(p=0.21)$. By selecting a group of `superpixels' from the input image, the classifier provides visual explanations to the top predicted labels.}
\label{fig:lime}
\end{figure*}

Consider \bm{$g \in G$}, the explanation as a model from a class of potentially interpretable models \bm{$G$}. Here, \bm{$g$} can be decision trees, linear models, or other models of varying interpretability. 
Let explanation complexity be measured by \bm{$\Omega(g)$}. 
If \bm{$\pi_x(z)$} is a proximity measure between two instances \bm{$x$} and $z$ around \bm{$x$}, and \bm{$\mathcal{L}(f, g, \pi_x)$} represents faithfulness of \bm{$g$} in approximating \bm{$f$} in locality defined by \bm{$\pi_x$}, then, explanation \bm{$\EModel$} for the input data sample \bm{$x$} is given by the LIME equation:

\begin{equation}\label{eq:limeeq}
\EModel(x) = \argmin_{g \in G}\;\;\mathcal{L}(f, g, \pi_x) + \Omega(g)
\end{equation}

Now, in Equation \ref{eq:limeeq}, the goal of LIME optimization is to minimize the locality-aware loss \bm{$\mathcal{L}(f, g, \pi_x)$} in a model agnostic way. Example visualization of LIME algorithm on a single instance is illustrated in Figure \ref{fig:lime}. Algorithm \ref{alg:limelocal} shows the steps to explain the model for a single input sample and the overall procedure of LIME. Here, for the input instance we permute data by finding a superpixel of information (`fake' data). Then, we calculate distance (similarity score) between permutations and original observations. Now, we know how different the class scores are for the original input and the new `fake' data.

We can then make predictions on new `fake' data using the complex model $f$. This depends on the amount of superpixels you choose from the original data. The most descriptive feature can be picked which improved prediction on the permuted data. If we fit a simple model, often times a locally weighted regression model, to the permuted data with \bm{$m$} features and similarity scores as weights, we can use the feature weights, or coefficients, from the simple model to make explanations for the local behavior of the complex model. Recent years have seen many research improving and extending the LIME algorithm to a variety of new tasks. We summarize a few of them below:
\begin{itemize}
    \item In \cite{Mishra2017Local}, Mishra et al. extended LIME algorithm to music content analysis by temporal segmentation, and frequency and time-frequency segmentation of input mel-spectogram. Their approach was called Sound-LIME (SLIME) and was applied to explain the predictions of a deep vocal detector. 
    \item In \cite{Peltola2018Local}, Tomi Peltola described a Kullback–Leibler divergence based LIME called KL-LIME to explain Bayesian predictive models. Similar to LIME, the explanations are generated using an interpretable model, whose parameters are found by minimizing the KL-divergence from the predictive model. Thus, local interpretable explanations are generated by projecting information from the predictive distribution to a simpler interpretable probabilistic explanation model.
    \item In \cite{Rehman2019Dlime}, Rehman et al. used agglomerative Hierarchical Clustering (HC) and K-Nearest Neighbor (KNN) algorithms to replace the random perturbation of the LIME algorithm. Here, authors use the HC method to group training data together as clusters and the KNN is used to find closest neighbors to a test instance. Once the KNN picks a cluster, that cluster is passed as the input data perturbation instead of a random perturbation as in LIME algorithm. Authors report that their approach generates model explanations which are more stable than traditional LIME algorithm.
    \item In \cite{Bramhall2020Qlime}, Bramhall et al. adjusted the linear relations of LIME to consider non-linear relationships using a quadratic approximation framework called Quadratic-LIME (QLIME). They achieve this by considering the linear approximations as tangentials steps within a complex function. Results on a global staffing company dataset suggests that the mean square loss (MSE) of LIME's linear relationship at local level improves while using QLIME. 
    \item In \cite{Shi2020Modified}, Shi et al. introduced a replacement method to pick superpixels of information for image data using Modified Perturbed Sampling operation for LIME (MPS-LIME). Authors converted the traditional superpixel picking operation into a clique set construction problem by converting the superpixels to an undirected graph. The clique operation improves the runtime due to a considerable reduction in the number of perturbed samples in the MPS-LIME method. Authors compared their method with LIME using Mean Absolute Error (MAE) and Coefficient of determination $R^2$ and reported better results in terms of understandability, fidelity, and efficiency.

\end{itemize}

\subsubsection{SHapley Additive exPlanations (SHAP)}
A game theoretically optimal solution using Shapley values for model explainability was proposed by Lundberg et al. \cite{Lundberg2017}. SHAP explains predictions of an input \bm{$x$} by computing individual feature contributions towards that output prediction. By formulating the data features as players in a coalition game, Shapley values can be computed to learn to distribute the payout fairly. 

In SHAP method, a data feature can be individual categories in tabular data or superpixel groups in images similar to LIME. SHAP then deduce the problem as a set of linear function of functions where the explanation is a linear function of features \cite{Molnar2019}. If we consider \bm{$g$} as the explanation model of an ML model \bm{$f$}, \bm{$z'\in\{0,1\}^M$} as the coalition vector, \bm{$M$} the maximum coalition size, and $\boldmath{\phi_j\in\mathbb{R}}$ the feature attribution for feature \bm{$j$}, \bm{$g(z')$} is the sum of bias and individual feature contributions such that:

\begin{equation} \label{eq:shap}
    g(z')=\phi_0+\sum_{j=1}^M\phi_jz_j'
\end{equation}

Lundberg et al. \cite{Lundberg2017} further describes several variations to the baseline SHAP method such as KernelSHAP which reduces evaluations required for large inputs on any ML model, LinearSHAP which estimates SHAP values from a linear model's weight coefficients given independent input features, Low-Order SHAP which is efficient for small maximum coalition size \bm{$M$}, and DeepSHAP which adapts DeepLIFT method \cite{Sundararajan2017} to leverage the compositional nature of deep neural networks to improve attributions. Since KernelSHAP is applicable to all machine learning algorithms, we describe it in Algorithm \ref{alg:kernalshap}. The general idea of KernelSHAP is to carry out an additive feature attribution method by randomly sampling coalitions by removing features from the input data and linearizing the model influence using SHAP kernels.

\begin{algorithm}[!b]
    \small
    \caption{KernelSHAP Algorithm for Local Explanations}
    \label{alg:kernalshap}
\begin{algorithmic}[1]
    \Statex \textbf{Input:} classifier \bm{$f$}, input sample \bm{$x$}
    \Statex \textbf{Output:} explainable coefficients from the linear model

    \State \bm{$z_k \leftarrow \text{SampleByRemovingFeature}(x)$}
    \State \bm{$z_k \leftarrow h_x(z_k) $} \Comment{$h_x$ is a feature transformation to reshape to $x$}
    \State \bm{$y_k \leftarrow f(z_k) $}
    \State \bm{$W_x \leftarrow SHAP(f,z_k, y_k)$}
    \State \bm{$\texttt{LinearModel(}W_x\texttt{).fit()}$}
    \State Return \texttt{LinearModel.coefficients()}
\end{algorithmic}
\end{algorithm}

SHAP was also explored widely by the research community, was applied directly, and improved in many aspects. Use of SHAP in the medical domain to explain clinical decision-making and some of the recent works which have significant merits are summarized here:

\begin{itemize}
    \item In \cite{Antwarg2019Explaining}, Antwarg et al. extended SHAP method to explain autoencoders used to detect anomalies. Authors classify anomalies using the autoencoder by comparing the actual data instance with the reconstructed output. Since the final output is a reconstruction, authors suggests that the explanations should be based on the reconstruction error. SHAP values are found for top performing features and were divided into those contributing to and offsetting anomalies.
    \item In \cite{Sundararajan2019Many}, Sundararajan et al. express various disadvantages of SHAP method such as generating counterintuitive explanations for cases where certain features are not important. This `uniqueness' property of attribution method is improved using Baseline Shapley (BShap) method. Authors further extend the method using Integrated Gradients to the continuous domain.
    \item In \cite{Aas2019Explaining}, Aas et al. explored the dependence between SHAP values by extending KernelSHAP method to handle dependent features. Authors also presented a method to cluster Shapley values corresponding to dependent features. A thorough comparison of the KernelSHAP method was carried out with four proposed methods to replace the conditional distributions of KernelSHAP method using empirical approach and either the Gaussian or the Gaussian copula approaches.
    \item In \cite{Lundberg2020Tree}, Lundberg et al. described an extension of SHAP method for trees under a framework called TreeExplainer to understand the global model structure using local explanations. Authors described an algorithm to compute local explanation for trees in polynomial time based on exact Shapley values. 
    \item In \cite{VegaGarcia2020}, VegaGarcia et al. describe a SHAP-based method to explain the predictions of time-series signals involving Long Short-Term Memory (LSTM) networks. Authors used DeepSHAP algorithm to explain individual instances in a test set based on the most important features from the training set. However, no changes in the SHAP method was done, and explanations were generated for each time step of each input instances.
\end{itemize}

\subsection{Global Explanations}\label{subsec:global}
AI model behavior for a suite of input data points could provide insights on the input features, patterns, and their output correlations thereby promoting transparency of model behavior.
Various globally explainable methods deduce the complex deep models to linear counterparts which are easier to interpret. 
Rule-based and tree-based models such as decision trees are inherently globally interpretable. Output decision of individual branches of a tree can be traced back to the source. Similarly, linear models are often fully explainable given model parameters. 

\begin{figure}[!b]
\begin{center}
\includegraphics[width=\linewidth]{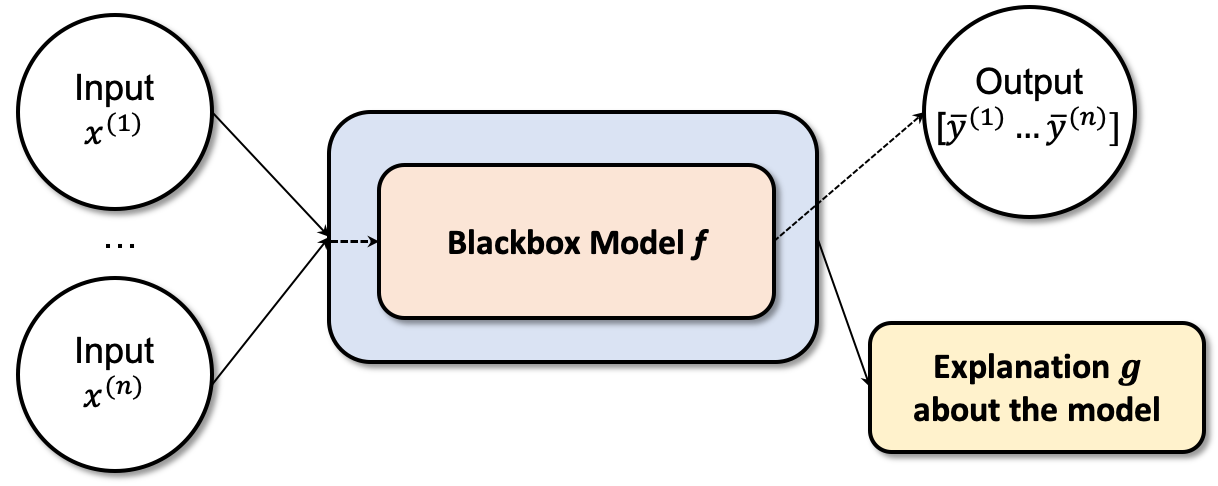}
\end{center}
  \caption{High-level illustration of a globally explainable algorithm design.}
\label{fig:globalmodel}
\end{figure}

Generally, globally explainable methods work on an array of inputs to summarize the overall behavior of the blackbox model as illustrated in Figure \ref{fig:globalmodel}. 
Here, the explanation \bm{$g_f$} describes the feature attributions of the model as a whole and not just for individual inputs. Thus, global explainability is important to understand the general behavior of the model \bm{$f$} on large distributions of input and previously unseen data.

\subsubsection{Global Surrogate Models}
Global surrogate models could be used as a way to approximate the predictions of highly non-linear AI models with an interpretable linear model or a decision tree. Global explanations answers the \textbf{'How'} in XAI, specifically ``How generalized is my AI model?", ``How do variations of my AI model perform?". A general use case of surrogate models in deep learning would be extraction of feature-rich layer embeddings for test inputs and training a linear classifier on the embeddings.
The coefficients of the linear model could give insights to how the model behaves. In a high-level, SHAP and LIME can both be considered as surrogate models with different methodology to understand the local correlations than linear models. SpRAy technique we will see in Section \ref{subsec:spray} also extract local features from a group of data to understand model behavior.

\subsubsection{Class Model Visualization}
Activation maximization \cite{Erhan2010} introduced in Section \ref{subsec:actmax} can be also expanded as a global method using Class Model Visualization as described by Simonyan et al. \cite{Simonyan2013}. Here, a given a trained ConvNet \bm{$f$} and a class of interest \bm{$c$}, the goal is to generate image visualizations \bm{$I'$} which is representative of $c$. This is based on the scoring methods used to train \bm{$f$} which maximizes the class probability score \bm{\bm{$S_c(I)$}} for \bm{$c$}, such that:

\begin{equation}
I^{'} = \arg \max _{I} S_{c}(I)-\lambda\|I\|_{2}^{2}
\end{equation}

\begin{figure}[!t]
\begin{center}
\includegraphics[width=\linewidth]{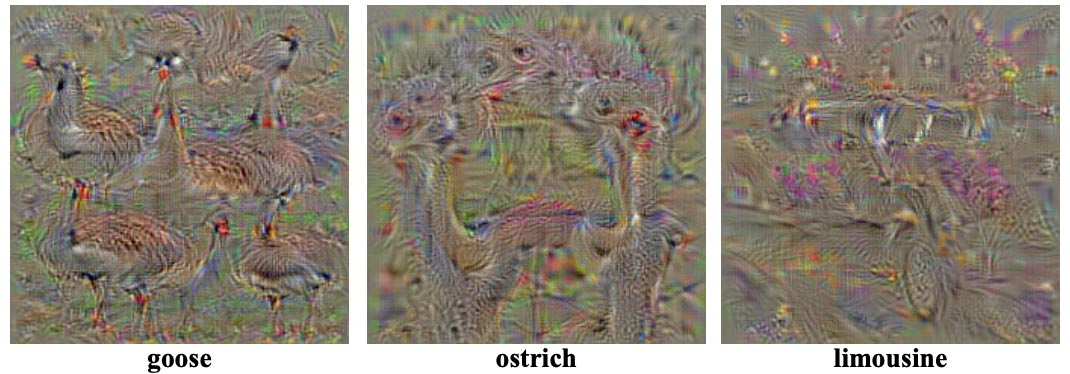}
\end{center}
  \caption{Numerically computed images from \cite{Simonyan2013} which uses the class-model visualization method to generate images representing the target class mentioned in the illustration.}
\label{fig:simonyan-class-model}
\end{figure}

Thus, the generated images provides insight to what the blackbox model had learnt for a particular class in the dataset. Images generated using this technique is often called `deep dream' due to the colorful artefacts generated in the visualizations corresponding to the output class under consideration. Figure \ref{fig:simonyan-class-model} illustrates three numerically computed class appearance models learnt by a CNN model for \textit{goose}, \textit{ostrich}, and \textit{limousine} classes respectively.
 
\subsubsection{LIME Algorithm for Global Explanations}
LIME model \cite{Ribeiro2016} was extended with a submodular pick algorithm (SP-LIME) to understand the global correlations of the model under study. This way, LIME provides a global understanding of the model from the individual data instances by providing a non redundant global decision boundary of the machine learning model. Generating global importance of individual features is done using a submodular pick algorithm (hence called SP-LIME). Algorithm \ref{alg:limeglobal} describes the steps to generate a global explanation to the blackbox model \bm{$f$} by learning individual feature importance of input samples \bm{$x_1, \ldots, x_n$ $\in X$}. 

\begin{algorithm}[!ht]
    \small
    \caption{LIME Algorithm for Global Explanations}
    \label{alg:limeglobal}
\begin{algorithmic}[1]
    \Statex \textbf{Input:} classifier \bm{$f$}, input samples \bm{$x_1, \ldots, x_n$ $\in X$}
    \Statex \textbf{Output:} explanation matrix after submodular pick

    \State \texttt{Define} instances \bm{$X$} and budget \bm{$B$}
    \For {\bm{$x \in X$}}
        \State \bm{$f_{LIME} \leftarrow LIME(f, x)$}
    \EndFor
    
    \State \texttt{Select} \bm{$B$} features from \bm{$f_{LIME}$}
    
    \Statex \texttt{Submodular Pick}: 
    \State \bm{$M \leftarrow \texttt{GenerateMatrix(} X, B \texttt{)}$}
    \State $X_{min} \leftarrow \texttt{GreedyOptimization(}M\texttt{)}$
    
\end{algorithmic}
\end{algorithm}

If $B$ is the number of explanations to inspect called Budget, $W$, the explanation matrix, we start with explaining all instances \bm{$x \in X$} using LIME algorithm explained in Section \ref{subsec:limelocal}. In the domain of images, \bm{$X$} represents individual input images and \bm{$B$} represent the number of superpixels selected for the LIME algorithm. Then, we select \bm{$B$} features from \bm{$f$} which represents the image better. The submodular pick algorithm starts by generating a matrix of size \bm{$X \times B$} and applying greedy optimization on the matrix such that it chooses minimum number of inputs \bm{$min(X)$} which covers the most number of features \bm{$max(F)$}. Here, SP-LIME works similar to a surrogate model by first extracting the independent explainability vectors using LIME operation. Hence, computational overhead, accuracy, and complexity depends partly on the amount out data used to understand the model globally.

\subsubsection{Concept Activation Vectors (CAVs)}\label{subsec:tcav}
In \cite{Kim2018Interpretability}, Kim et al. introduced Concept Activation Vectors (CAVs), a global explainability method to interpret the internal states of a neural network in human-friendly concept domain. Here, if we consider the machine learning model \bm{$f(.)$} as a vector space \bm{$E_m$} spanned by basis vector \bm{$e_m$}, we see that human understanding can be modelled as vector space \bm{$E_h$} and implicit vectors \bm{$e_h$} which correspond to human-understandable concepts \bm{$C$}. Hence, the explanation function of the model in a global sense, \bm{$g$}, becomes \bm{$g: E_m \rightarrow E_h$}.

\begin{figure*}[ht]
\begin{center}
\includegraphics[width=0.9\linewidth]{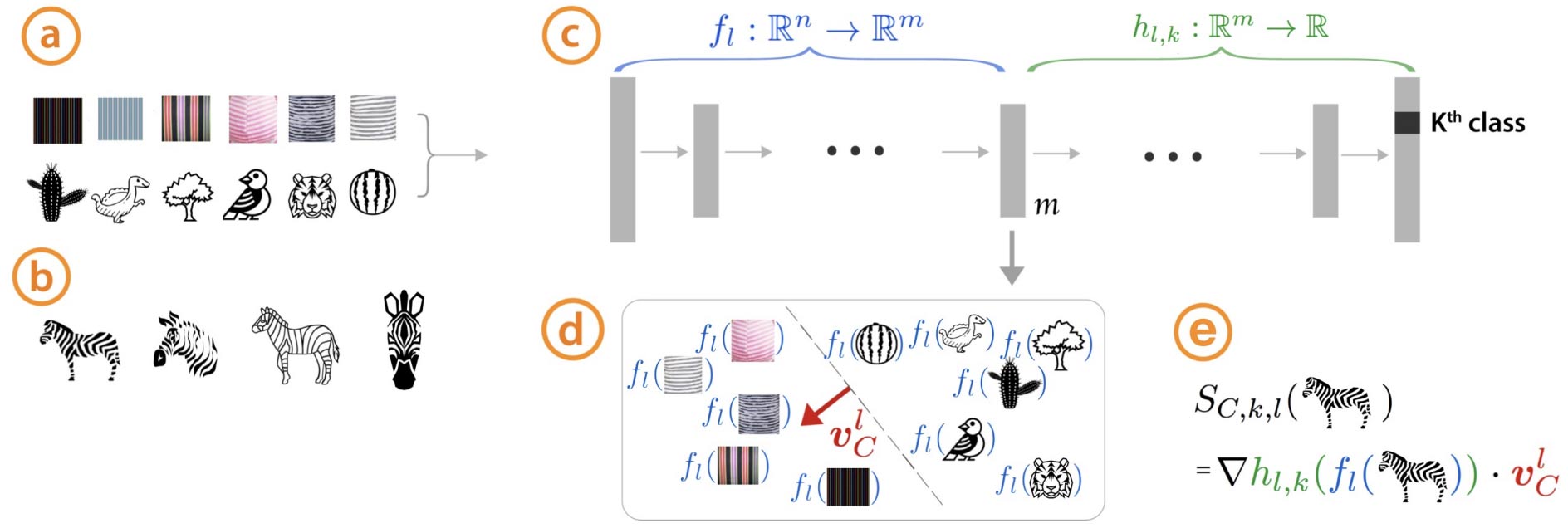}
\end{center}
  \caption{Figure illustrates the TCAV process \cite{Kim2018Interpretability} where (a) describe random concepts and examples, (b) labelled examples from training data, (c) trained neural network, (d) linear model segregating the activations extracted from specific layers in the neural network for the concepts and random examples, and (e) finding conceptual sensitivity using directional derivatives.}
\label{fig:tcav}
\end{figure*}

Now, human understandable concepts are generated from either input features of training data or user-provided data to simplify the lower-level features of the input domain. For example, a zebra can be deduced to positive concepts \bm{$P_C$} such as stripes as illustrated in Figure \ref{fig:tcav}. A negative set of concepts, $N$, can be gathered, for example a set of random photos, to contrast the concepts for zebra. Layer activations for layer \bm{$j$} of \bm{$f$}, \bm{$z_j$} is calculated for both positive and negative concepts. The set of activations are trained using a binary classifier to distinguish between: 
\bm{$\{f_j(\bm{x}) : \bm{x} \in P_C\}$} and
\bm{$\{f_j(\bm{x}) : \bm{x} \in N\}$}.

Authors proposed a new method, Testing with CAVs (TCAV), which uses directional derivatives similar to gradient based methods to evaluate the sensitivity of class predictions of \bm{$f$} to the changes in given inputs towards the direction of the concept \bm{$C$} for a specific layer \bm{$j$}. If \bm{$h(j,k)$} is the logit of layer \bm{$j$} for class \bm{$k$} for a particular input, conceptual sensitivity of a class \bm{$k$} to \bm{$C$} can be computed as directional derivative \bm{$S_{C,k,j}(\bm{x})$} for a concept vector \bm{$\bm{v}_C^j \in \mathbb{R}^m$}:

\begin{eqnarray}
S_{C,k,j}(\bm{x}) & = &
\lim\limits_{\epsilon \rightarrow 0} \frac{h_{j, k}(z_j(\bm{x}) + \epsilon \bm{v}_C^j) - h_{j,k}(z_j(\bm{x}))}{\epsilon} 
\nonumber \\
& = & \nabla h_{j,k}(z_j(\bm{x})) \cdot \bm{v}_C^j ,
\vspace*{-2ex}
\end{eqnarray}

A TCAV score can be calculated to find the influence of inputs towards \bm{$C$}. If \bm{$X_k$} denotes all inputs with label \bm{$k$}, TCAV score is given by:

\begin{equation}
 TCAV_{C, k, j} = \frac{\left\lvert \left\{ 
                    \bm{x}\in X_k : 
                    S_{C,k,j}(\bm{x}) > 0 
                    \right\} \right\rvert}%
                   {\left\lvert X_k \right\rvert}
\end{equation}

TCAV unfortunately could generate meaningless CAVs if the input concepts are not picked properly. For example, input concepts generated randomly would inherently generate bad linear models for binary classification and thus TCAV score wouldn't be a good identifier for global explainability. Also, concepts with high correlations or shared objects in the data, such as cars and roads, could decrease the efficiency of TCAV method. Human bias in picking the concepts also is a considerable disadvantage of using concepts for explainability. The CAV method was further improved in numerous research papers which involved the primary author of CAV \cite{Kim2018Interpretability}:

\begin{itemize}
    \item In \cite{Ghorbani2019Towards}, Ghorbani et al. described a method called  Automatic Concept-based Explanations (ACE) to globally explain a trained classifier without human supervision unlike TCAV method. Here, authors carry out a multi-resolution segmentation of instance to be explained. This generates multiple resolution segments from the same class. All segments are reshaped to similar input sizes and activations of each segment is found with respect to a specific chosen bottleneck layer. Clustering the activations and removing outliers reveals similarities within activations. 
    
    TCAV scores of individual concepts provide an importance score of the same for particular classification. Authors carried out human subject experiments to evaluate their method and found inspiring results. One research question that arise is the importance of clusters in decision-making. Authors showed that, by stitching the clustered concepts together as an image, a trained InceptionV3 deep neural network was capable of classifying the stitched image as the correct class category. This tends to show that the extracted concepts are suitable for decision-making within the deep learning model.

    \item Work done by Goyal et al. \cite{Goyal2019Explaining} improved TCAV method by proposing a Causal Concept Effect (CaCE) model which looks at the causal effect of presence or absence of high-level concepts towards deep learning model's prediction. Methods such as TCAVs can suffer from confounding of concepts which could happen if the training data instances have multiple classes in them, even with low correlation between the classes. Also, biases in dataset could influence concepts, as well as colors in the input data.
    
    CaCE can be computed exactly if we can change concepts of interest by intervening the counterfactual data generation. Authors call this Ground truth CaCE (GT-CaCE) and also elaborate a way to estimate CaCE using Variational Auto Encoders (VAEs) called VAE-CaCE. Experimental results on four datasets suggest improved clustering and performance of the CaCE method even when there are biases or correlations in the dataset.
    
    \item In \cite{Yeh2019Concept}, Yeh et al. introduced ConceptSHAP to define an importance or ``completeness" score for each discovered concept. Similar to ACE method mentioned earlier, one of the aims of ConceptSHAP is to have concepts consistently clustered to certain coherent spatial regions. However, ConceptSHAP finds the importance of each individual concepts with high completeness score from a set of $m$ concept vectors \bm{$C_s = \{c_1, c_2, \ldots, c_m\}$} by utilizing Shapley values for importance attribution.
\end{itemize}

\subsubsection{Spectral Relevance Analysis (SpRAy)}
\label{subsec:spray}
SpRAy technique by Lapuschkin et al. \cite{Lapuschkin2019} builds on top of the local instance based LRP explanations. In specific, authors described a spectral clustering algorithm on local explanations provided by LRP to understand the decision-making process of the model globally. By analyzing the spatial structure of frequently occurring attributions in LRP instances, SpRAy identifies normal and abnormal behavior of machine learning models.

Algorithm \ref{alg:spary} explains the SpRAy technique in detail. We start by finding local relevance map explanations to every individual data instances $x \in X$ using LRP method. The relevance maps are downsized to uniform shape and size to improve computation overhead and generate tractable solutions. Spectral cluster analysis (SC) is carried out on the LRP attribution relevance maps to cluster the local explanations in a high-dimensional space. An eigenmap analysis is carried out to find relevant clusters by finding the eigengap (difference in two eigenvalues) of successive clusters. After completion, important clusters are returned to users. The clusters can be optionally visualized using t-Stochastic Neighbor Embedding (t-SNE) visualizations.

\begin{algorithm}[ht]
    \small
    \caption{SpRAy Analysis Algorithm on LRP Attributions}
    \label{alg:spary}
\begin{algorithmic}[1]
    \Statex \textbf{Input:} classifier \bm{$f$}, input samples \bm{$x^{(1)}, \ldots, x^{(n)}$}
    \Statex \textbf{Output:} clustered input samples
    \For {\bm{$x^{(i)} \in X$}}
        \State \bm{$f_{SpRAy} \leftarrow LRP(f, x^{(i)})$}
    \EndFor
    \State \texttt{Reshape} \bm{$f_{SpRAy}$}
    \State \bm{$clusters \leftarrow SC(f_{SpRAy})$}
    \State \bm{$clusters^* \leftarrow \texttt{EigenMapAnalysis(} clusters \texttt{)}$ }
    \State \texttt{Return} \bm{$clusters^*$}
    \State \texttt{Optional: Visualize t-SNE(}\bm{$clusters^*$}\texttt{)}
\end{algorithmic}
\end{algorithm}

\subsubsection{Global Attribution Mapping}
When features have well defined semantics, we can treat attributions as weighted conjoined rankings \cite{Ibrahim2019Global} with each feature as a rank vector \bm{$\sigma$}. After finding local attributions, global attribution mapping finds a pair-wise rank distance matrix and cluster the attribution by minimizing cost function of cluster distances. This way, global attribution mapping can identify differences in explanations among subpopulations within the clusters which can trace the explanations to individual samples with tunable subpopulation granularity.

\subsubsection{Neural Additive Models (NAMs)}\label{subsec:nam}

In \cite{Agarwal2020Neural}, Agarwal et al. introduced a novel method to train multiple deep neural networks in an additive fashion such that each neural network attend to a single input feature. Built as an extension to generalized additive models (GAM), NAM instead use deep learning based neural networks to learn non-linear patterns and feature jumping which traditional tree-based GAMs cannot learn. NAMs improved accurate GAMs introduced in \cite{Caruana2015} and are scalable during training to several GPUs. 

\begin{figure}[!ht]
\begin{center}
\includegraphics[width=0.9\linewidth]{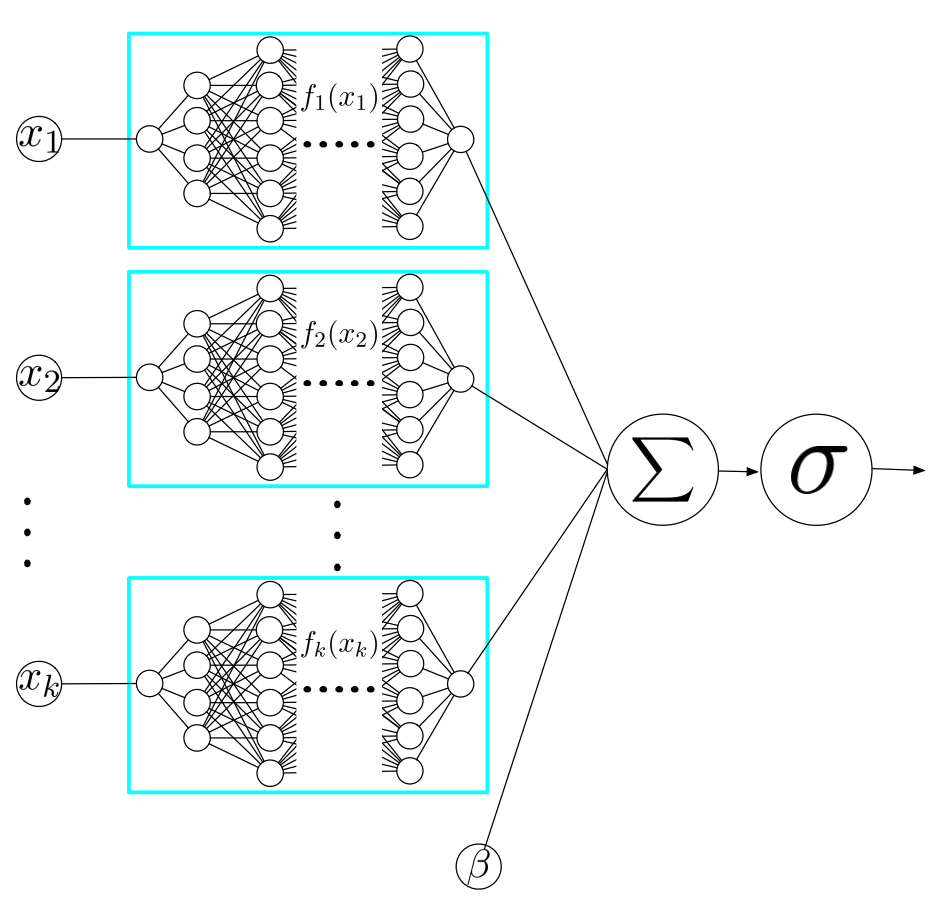}
\end{center}
  \caption{A high-level diagram of the interpretable NAM architecture for binary classification is illustrated \cite{Agarwal2020Neural}. Functions $f_i$ is used to learn from corresponding. individual features in $x_i$.}
\label{fig:nam}
\end{figure}

Consider a general GAM of the form:

\begin{equation}\label{eq:GAM}
g(\mathbb{E}[y])=\beta+f_{1}\left(x_{1}\right)+f_{2}\left(x_{2}\right)+\cdots+f_{K}\left(x_{K}\right)
\end{equation}

where \bm{$f_i$} is a univariate shape function with \bm{$\mathbb{E}[f_i] = 0$}, \bm{$x \in x_1, x_2, \ldots, x_K$} is the input with K features, \bm{$y$} is the target variable, and \bm{$g(.)$} is a link function. NAMs can be generalized by parameterizing the functions \bm{$f_i$} with neural networks with several hidden layers and neurons in each layer. We can see individual neural networks applied to each features $x_i$. The outputs of each $f_i$ is combined together using a summing operation before applying an activation. A high-level diagram of NAM is provided in \ref{fig:nam} taken from the source paper. 

Authors proposed exp-centered (ExU) hidden units to overcome the failure of ReLU activated neural networks with standard initializations to fit jagged functions. NAMs should be able to learn jagged functions due to sharp changes in features in real-world datasets often encountered in GAMs. For ExU hidden units, the unit function can be calculated as \bm{$h(x) = f(e^w*(x-b))$}, where \bm{$x$}, \bm{$w$}, and \bm{$b$} are the inputs, weights, and biases parameters. Authors used a weight initialization of training from a normal distribution \bm{$\mathcal{N}(x, 0.5)$} with \bm{$x \in[3,4]$}. This globally explainable model provides average score of shape functions of individual neural networks to provide interpretable contributions of each features as positive and negative values. Negative values reduce the class probability while positive values improve the same.

NAM is an interesting architecture because we can generate exact explanations of each feature space with respect of an output prediction. Newer research could open up venues to expand the ideas to CNNs and for other domains such as text.

\section{Differences in the Methodology}
\label{sec:methodology}
Based on the core algorithmic approach followed in the XAI method, we can categorize XAI methods as the ones which focus on the changes or modifications input data and the ones which focus on the model architecture and parameters. These fundamental changes are categorized in our survey as perturbation-based and backpropagation-based respectively.

\subsection{Perturbation-Based}
Explanations generated by iteratively probing a trained machine learning model with different variations of the inputs generally fall under perturbation based XAI techniques. These perturbations can be on a feature level by replacing certain features by zero or random counterfactual instances, picking one or group of pixels (superpixels) for explanation, blurring, shifting, or masking operations, etc. As we discussed in the prior sections, LIME algorithm works on superpixels of information or features as illustrated in Figure \ref{fig:lime}. By iteratively providing input patches, visual explanations of individual superpixels are generated. SHAP has a similar method of probing feature correlations by removing features in a game theoretic framework. Intuitively, we see that methods trying to understand neuronal activities and the impact of individual features to a corresponding class output by any input perturbations mentioned above can be categorized as a group of method, which we here call \textbf{perturbation-based} XAI method. The methods described in this section are further summarized in Table \ref{tab:perturbationmethods}.

\begin{table*}[ht]
\caption{Summary of published research in perturbation-based methods}
\label{tab:perturbationmethods}
\centering

\begin{tabular}{||p{3cm}|p{5.5cm}|p{2cm}|p{5cm}||}
\hline
Method Name 
    & Interpretation Perspective 
        & Applied Network 
            & Comments and Discussions \\ 
\hline\hline
DeConv Nets by Zeiler et al. \cite{Zeiler2014}  
    & Neural activation of individual layers by occluding input instance and visualizing using DeConv Nets 
        & AlexNet 
            & Authors trained an AlexNet model on ImageNet dataset and layer-wise filter visualizations were carried out, studied feature generalization, and brought  important insights in dataset bias and issues with small training samples. \\
\hline
LIME by Ribeiro et al. \cite{Ribeiro2016}       
    & Iterative perturbation to input data instance by finding superpixels 
        & - 
            & Authors generated locally faithful explanations using input perturbations around a point of interest. A human/user study was carried out to assess the impact of using LIME as an explanation and found that explanations can improve a untrustworthy classifier. \\
\hline
SHAP by Lundberg et al. \cite{Lundberg2017}       
    & Probing feature correlations by removing features in a game theoretic framework 
        & - 
            & SHAP produced consistently better results than LIME. A user study indicated that SHAP explanations are consistent with human explanations. However, as we will see in the evaluation section, some recent studies argue that SHAP values, albeit good in generating explanations, does not improve final decision making.\\
\hline
Prediction Difference Analysis by Zintgraf et al. \cite{Zintgraf2017}
    & By studying $f$ removing individual features from $x$, find the positive and negative correlation of individual features towards the output. 
        & AlexNet, GoogLeNet, VGG 
            & One of the first works to look at positive and negative correlation of individual features towards the output by finding a relevance value to each input feature. Trained various models on ImageNet dataset to understand the support for the output classes from various layers of deep nets.\\
\hline
Randomized Input Sampling for Explanation by Petsiuk et al. \cite{Petsiuk2018}
    & Study saliency maps by randomized masking of inputs 
        & ResNet50, VGG16 
            &  - \\
\hline
Randomization and Feature Testing by Burns et al. \cite{Burns2019}
    & Counterfactual replacements of features to study feature importance 
        & Inception V3, BERT 
            &  - \\
\hline

\end{tabular}
\end{table*}

\subsubsection{DeConvolution nets for Convolution Visualizations}
\begin{figure}[!b]
\begin{center}
\includegraphics[width=\linewidth]{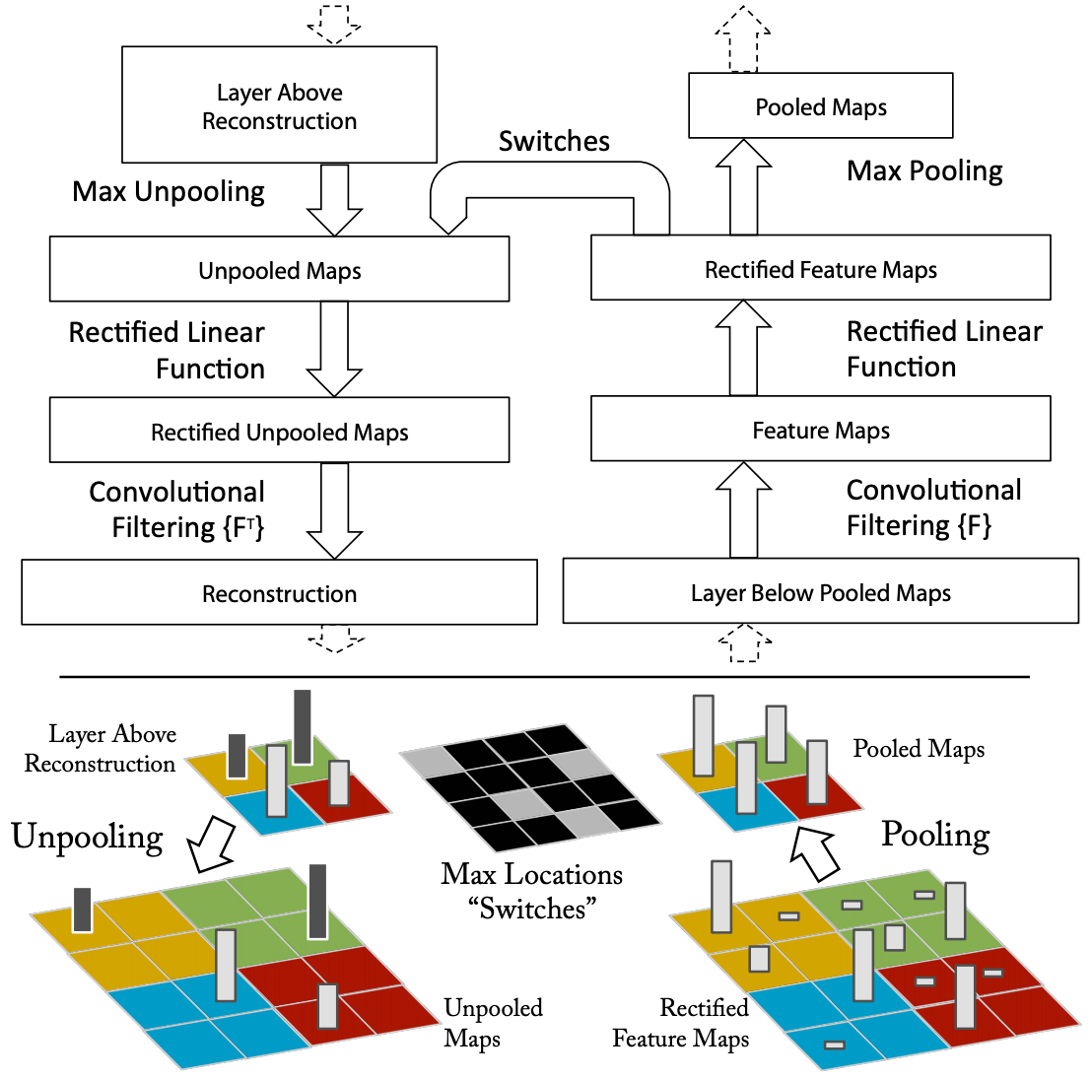}
\end{center}
  \caption{Deconvolution operation is applied using a DeConv layer attached to the end of a ConvNet. Here the DeConvNet generates an approximate version of the convolution features thereby providing visual explanations. Figure from \cite{Zeiler2014}.}
\label{fig:deconvnet}
\end{figure}

Zeiler et al. \cite{Zeiler2014} visualized the neural activations of individual layers of a deep convolutional network by occluding different segments of the input image and generating visualizations using a deconvolution network (DeConvNet). DeConvNets are CNNs designed with filters and unpooling operations to render opposite results than a traditional CNN. Hence, instead of reducing the feature dimensions, a DeConvNet, as illustrated in Figure \ref{fig:deconvnet}, is used to create an activation map which maps back to the input pixel space thereby creating a visualization of the neural (feature) activity. The individual activation maps could help understand what and how the internal layers of the deep model of interest is learning - allowing for a granular study of DNNs.

\subsubsection{Prediction Difference Analysis}
A conditional sampling based multi-variate approached was used by Zintgraf et al. \cite{Zintgraf2017} to generate more targeted explanations on image classification CNNs. By assigning a relevance value to each input features with respect to the predicted class $c$, the authors summarize the positive and negative correlation of individual data features to a particular model decision. Given an input feature $x$, its feature relevance can be estimated by studying the changes in model output prediction for the inputs with different hidden features. Hence, if $\xbacki$ denotes the set of all input features except $x$, the task is to find the difference between $p(c|\mathbf{x})$ and $p(c|\xbacki)$.

\subsubsection{Randomized Input Sampling for Explanation (RISE)}
\begin{figure}[!ht]
\begin{center}
\includegraphics[width=\linewidth]{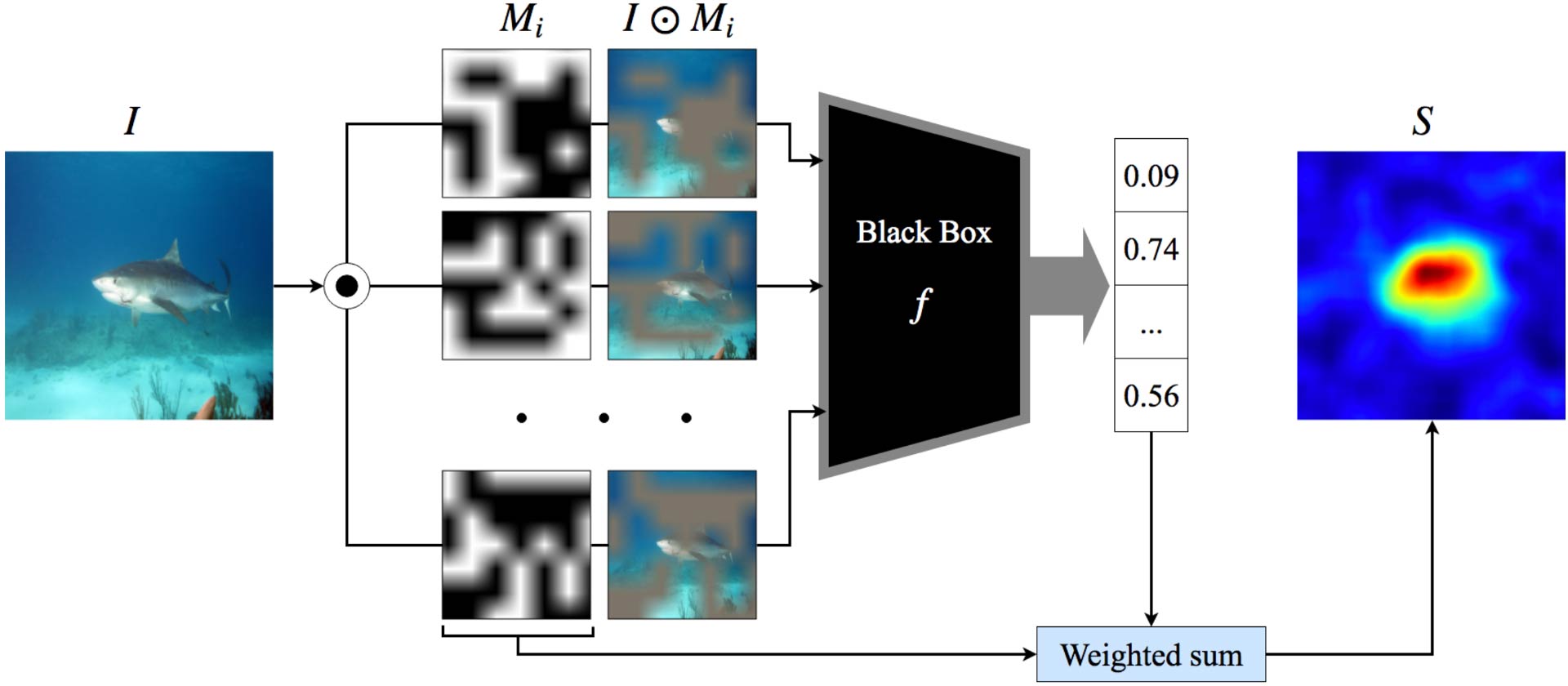}
\end{center}
  \caption{The input image given to a deep learning model is perturbed using various randomized masks. A confidence score is found out for individual masked inputs. A final saliency map is generated using a weighting function \cite{Petsiuk2018}.}
\label{fig:RISE}
\end{figure}

The RISE method introduced by Petsiuk et al. \cite{Petsiuk2018} perturb an input image by multiplying it with randomized masks. The masked images are given as inputs and the saliency maps corresponding to individual images are captured. Weighted average of the masks according to the confident scores is used to find the final saliency map with a positive valued heatmap for individual predictions. Importance maps of the blackbox prediction is estimated using Monte Carlo sampling.  
A high-level architecture is illustrated in Figure \ref{fig:RISE}.

\subsubsection{Randomization and Feature Testing}
The Interpretability Randomization Test (IRT) and the One-Shot Feature Test (OSFT) introduced by Burns et al. \cite{Burns2019} focuses on discovering important features by replacing the features with uninformative counterfactuals. Modeling the feature replacement with a hypothesis testing framework, the authors illustrate an interesting way to examine contextual importance. Unfortunately, for deep learning algorithms, removing one or more features from the input isn't possible due to strict input dimensions for a pre-trained deep model. Zero-ing out values or filling in counterfactual values might lead to unsatisfactory performance due to correlation between features.


\subsection{BackPropagation- or Gradient-Based}
Perturbation-based methods, as we saw in the previous section, focuses on variations in the input feature space to explain individual feature attributions of $f$ towards the output class $c$. Gradient-based explainability methods, in contrast, utilize the backward pass of information flow in a neural network to understand neuronal influence and relevance of the input $x$ towards the output. As we will see in the following subsections, majority of gradient-based methods focuses on either visualization of activations of individual neurons with high influence or overall feature attributions reshaped to the input dimensions. A natural advantage of gradient-based XAI methods are the generation of human understandable visual explanations.

\subsubsection{Saliency Maps}
As mentioned in sub-section \ref{subsec:saliencymap}, Simonyan et al. \cite{Simonyan2013} introduced a gradient based method to generate saliency maps for convolutional nets. DeConvNet work by Zeiler et al. \cite{Zeiler2014} mentioned previously as a perturbation method uses backpropagation for activation visualizations. DeConvNet work was impressive due to relative importance given to gradient value during backprop. With Rectified Linear Unit (ReLU) activation, a backprop on traditional CNNs would result in zero values for negative gradients. However, in DeConvNets, the gradient value is not clipped at zero. This allowed for accurate visualizations. Guided backpropagation methods \cite{Springenberg2015,Mahendran2016} are also another class of gradient based explanation which improved upon \cite{Simonyan2013}.

\subsubsection{Gradient class activation mapping (CAM)}
Most saliency methods use global average pooling layer for all pooling operations instead of maxpooling. Zhou et al. \cite{Zhou2016Learning} modified global average pooling function with class activation mapping (CAM) to localize class-specific image regions on an input image with a single forward-pass. Grad-CAM \cite{Selvaraju2017} and Grad-CAM++ \cite{Chattopadhay2018} improved the CAM operation for deeper CNNs and better visualizations.

\begin{figure}[!t]
\begin{center}
\includegraphics[width=\linewidth]{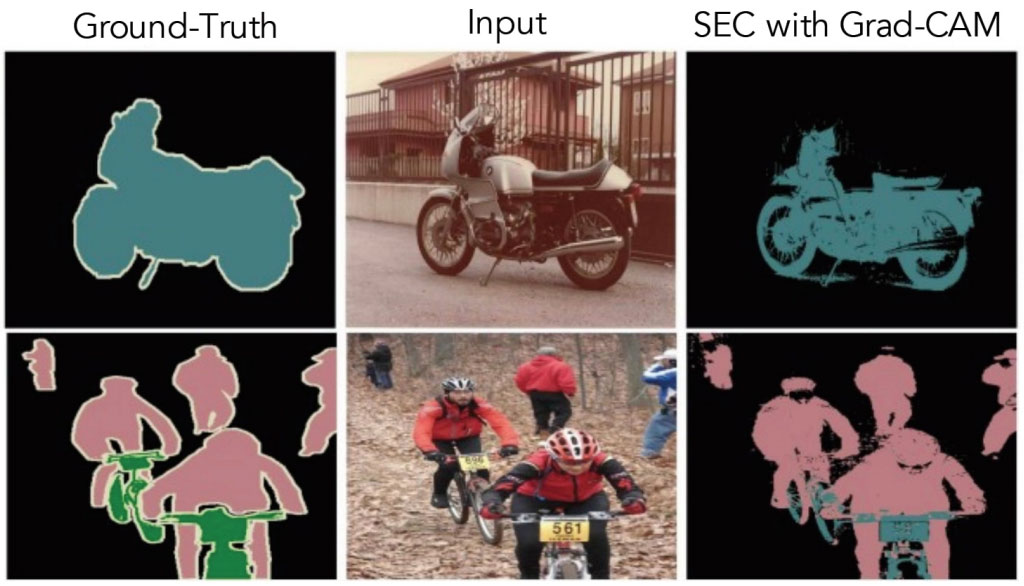}
\end{center}
  \caption{Illustration from \cite{Selvaraju2017} showing segmentation results by using Grad-CAM output as a seed.}
\label{fig:gradcam_segmentation}
\end{figure}

GradCAM is a class-discriminative attribution technique for localizing the neuronal activity of a CNN network. It allows class-specific query of an input image and also counterfactual explanations which highlights regions in the image which negatively contribute to a particular model output. GradCAM is successfully applied to explain classifiers in image classification, image segmentation, visual question answering (VQA), etc. Figure \ref{fig:gradcam_segmentation} illustrates a segmentation method utilizing GradCAM to improve the segmentation algorithm. Here, we see another example of using XAI explanations to improve performance of deep neural networks.

\subsubsection{Salient Relevance (SR) Maps}
Li et al. \cite{Li2019} proposed Salient Relevance (SR) map which is a context aware salience map based on the LRP of input image. Hence, the first step is to find LRP relevance map for input image of interest with the same input dimensions. A context aware salience relevance map algorithm takes the LRP relevance maps and finds a saliency value for individual pixels. Here, a pixel is salient if a group of neighboring pixels are distinct and different from other pixel patches in the same and multiple scales. This is done to differentiate between background and foreground layers of the image. 

To aid visualization, a canny-edge based detector is superimposed with the SR map to provide context to the explanation. We place SR in gradient based methods due to the use of LRP. Other relevance propagation methods based on Taylor decomposition \cite{Montavon2017} are also explored in literature, which are slightly different in the methodology but have the same global idea.

Algorithm \ref{alg:sr} describes the SR map generation in detail. Similar to SpRAy technique, we start with the LRP of the input instance. In contrast, we only find LRP attribution relevance score for a single input of interest $x$. Then a context aware saliency relevance (SR) map is generated by finding a dissimilarity measure based on the euclidean distance in color space and position. Multi-scale saliency at scales $r, \frac{r}{2}, \frac{r}{4}$ are found out and the immediate context of image $x$ based on an attention function is added to generate the SR map.

\begin{algorithm}[ht]
  \caption{Salient Relevance (SR) Algorithm}
  \label{alg:sr}
\begin{algorithmic}[1]
    \Statex \textbf{Input:} classifier $f$, input sample $x$, scale factor $r$
    \Statex \textbf{Output:} relevance map
    
    \State \bm{$f_{LRP} \leftarrow LRP(x)$}
    \State \bm{$\texttt{GenerateSRMap(}x, f_{LRP}\texttt{)}$}
    \State \bm{$S \leftarrow \texttt{MultiScaleSaliency(} r, \frac{r}{2}, \frac{r}{4} \texttt{)}$}
    \State \bm{$SRMap \leftarrow \texttt{AttentionFunction(} x, S\texttt{)}$}
    \State \texttt{Return} \bm{$SRMap$}
\end{algorithmic}
\end{algorithm}

\begin{table*}[!t]
\caption{Summary of published research in gradient-based methods}
\label{tab:gradientmethods}
\centering

\begin{tabular}{||p{3cm}|p{5.5cm}|p{2cm}|p{5cm}||}
\hline
Method Name 
    & Interpretation Perspective 
        & Applied Network 
            & Comments and Discussions \\ 
\hline\hline
Saliency Maps \cite{Simonyan2013, Zeiler2014, Springenberg2015, Mahendran2016} 
    & Visualizing gradients, neural activation of individual layers using DeConv nets, guided backpropagation, etc. as images. 
        & AlexNet, GoogLeNet, ResNet18, VGG16 
            &  A group of techniques which kicked-off gradient-based XAI research. As we will see in the evaluation section, these methods have serious disadvantages which needs to be improved.\\

\hline
Grad-CAM by Selvaraju et al. \cite{Selvaraju2017}
    & Localize neuronal activity flowing to last convolutional layer of a CNN to allow class-specific query with counterfactual explanations describing negative influence of input features as well. 
        & AlexNet, VGG16, ResNet, and more.  
            &  -\\
\hline
Salient Relevance by Li et al. \cite{Li2019}
    & Takes the LRP relevance maps and finds a saliency value for individual pixels. 
        & - 
            &  -\\
\hline
Axiomatic Attribution Maps by Sundararajan et al. \cite{Sundararajan2017}
    & Feature importance based on distance from a baseline instance 
        & GoogLeNet, LSTM based NMT, and more. 
            &  Introduced axioms or desirable qualities for gradient-based methods. Improved the saliency maps and gradient times input maps.\\
\hline
PatternNet and PatternAttribution by Kindermans et al. \cite{Kindermans2017Learning}
    & LRP-based method with back-projection of estimated signals to input space. Cleaner attributions based using root point selection algorithm. 
        & VGG16 
            &  - \\
\hline

\end{tabular}
\end{table*}

\subsubsection{Attribution Maps}
In \cite{Ancona2017}, Ancona et al. shows that the gradient method, where the gradient of output corresponding to input is multiplied by the input, is useful in generating an interpretable explanation to model outcomes. However, in \cite{Sundararajan2017}, authors proposed Integrated Gradients (IG) and argue that most gradient based lack in certain `axioms' which are desirable characteristics of any gradient based technique. Authors argue that methods such as DeepLift \cite{Shrikumar2017}, Layer-wise relevance propagation (LRP) \cite{Bach2015}, Deconvolutional networks (DeConvNets) \cite{Zeiler2014}, and Guided back-propagation \cite{Springenberg2015} have specific back-propagation logic that violates some axioms.

For each input data instance $x$, if we consider a baseline instance $x^{'} \in \mathbb{R}^{n}$, the attributions of $x$ on model $f$ can be summarized by computing the integral of gradients at all points of a straight-line path from baseline $x^{'}$ to $x$. This method is called the Integrated Gradients such that:

\begin{equation} \label{eq:intgrad}
\small
\text {IG}_j(x, x^{\prime}) := (x_{j}-x^{\prime}_{j})\times\int_{\alpha=0}^{1} \tfrac{\partial F(x^{\prime} + \alpha\times(x-x^{\prime}))}{\partial x_{j}  }~d\alpha
\end{equation}

where $j$ describes the dimension along which the gradient is calculated. 
During calculation in computers, the integral in equation \ref{eq:intgrad} is efficiently approximated using summation instead.
In many cases, baseline instance $x^{'}_i$ is chosen as a zero matrix or vector. For example, for image domain, the baseline image is chosen as a black image by default. For text classification, the baseline is a zero valued vector. However, choosing baselines arbitrarily could cause issues downstream. For example, a black baseline image could cause the attribution method to diminish the importance of black pixels in the source image.

Attribution prior \cite{Erion2019Learning} concept tries to regularize the feature attributions during model training to encode domain knowledge. A new method, Expected Gradients (EG) was also introduced in the paper as a substitute feature attribution method instead of Integrated Gradients. Together, the attribution prior and EG methods encodes prior knowledge from the domain to aid training process leading to better model interpretability. Equation \ref{eq:expgrad} shows how authors remove the influence of baseline images from integrated gradients by still following all the axioms of Integrated Gradient method. Here, $D$ is the distribution of underlying data domain.

\begin{equation} \label{eq:expgrad}
\small
\begin{split}
\text {EG}(x):=\int_{x^{\prime}}\left(\left(x_{j}-x_{j}^{\prime}\right) \int_{\alpha=0}^{1} \tfrac{\delta f\left(x^{\prime}+\alpha \times\left(x-x^{\prime}\right)\right)}{\delta x_j} \delta \alpha\right)\\ . p_{D}\left(x^{\prime}\right) \delta x^{\prime}
\end{split}
\end{equation}

Since an integration over the whole training distribution is intractable, authors proposed to reformulate the integral as expectations such that:

\begin{equation} \label{eq:expgrad2}
\small
\begin{split}
\text { EG}(x):=\underset{x^{\prime} \sim D, \alpha \sim U(0,1)}{\mathbb{E}}\left[\left(x_{j}-x_{j}^{\prime}\right) \tfrac{\delta f\left(x^{\prime}+\alpha \times\left(x-x^{\prime}\right)\right)}{\delta x_{j}}\right]
\end{split}
\end{equation}

\subsubsection{Desiderata of Gradient-based Methods}

Gradient-based methods, as we saw, mainly use saliency maps, class activation maps, or other gradient maps for visualization of important features. Recent research have found numerous limitations in gradient-based methods. To improve gradient-based XAI techniques, Sundararajan et al. \cite{Sundararajan2017} describes four desirable qualities (axioms) that a gradient based method needs to follow:

\begin{enumerate}
    \item \textbf{Sensitivity: } \textit{If for every input and baseline that differ in one feature but have different predictions then the differing feature should be given a non-zero attribution \cite{Sundararajan2017}}. For simple functions such as $f(x) = 1 - ReLU(1-x)$, the function value saturates for $x$ values greater than or equal to one. Hence, if we take simple gradients as an attribution method, sensitivity won't hold.
    
    \item \textbf{Implementation invariance: } \textit{Two networks are \emph{functionally equivalent} if their outputs are equal for all inputs, despite having very different implementations. Attribution methods should satisfy \emph{Implementation Invariance}, i.e., the attributions are always identical for two functionally equivalent networks \cite{Sundararajan2017}.} Methods such as DeepLift and LRP break implementation invariance because they use discrete gradients, and chain rule doesn't old for discrete gradients in general. Generally, if the model fails to provide implementation invariance, the attributions are potentially sensitive to unimportant features and aspects of the model definition.
    
    \item \textbf{Completeness: } Attributions should add up to the difference between output of model function $f$ for the input image $x$ and another baseline image $x^{'}$. $\Sigma_{i=1}^n Gradients_i(x) = f(x) - f(x^{'})$.

    \item \textbf{Linearity: } For a linearly composed neural network model $f_3$ which is a linear combination of two neural network models $f_1$ and $f_2$ such that $f_3 = a \times f_1 + b \times f_2$, then the attributions of the $f_3$ is expected to be a weighted sum of attributions for $f_1$ and $f_2$ with weights $a$ and $b$ respectively.
\end{enumerate}

Despite human understandable explanations, gradient-based explanation maps have practical disadvantages and raises various concerns in mission-critical applications. We explain some of these concerns in later sections.

\section{Model Usage or Implementation Level}
\label{sec:usage}
\begin{figure}[!b]
\begin{center}
\includegraphics[width=\linewidth]{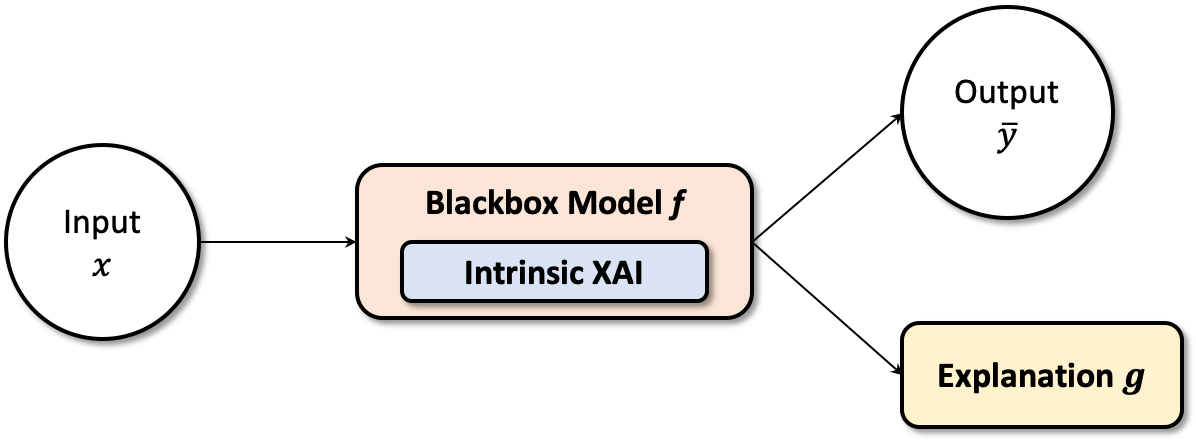}
\end{center}
  \caption{High-level illustration of model intrinsic explainability algorithms. Here, the explainability is baked into \bm{$f$} itself such that \bm{$f$} is naturally explainable.}
\label{fig:intrinsicmodel}
\end{figure}

\subsection{Model Intrinsic}
On a usage or implementation level, model intrinsic explainable methods have interpretable elements baked into them. These models are inherently interpretable either by following strict axioms, rule-based final decisions, granular explanations for decisions, etc. By definition, intrinsic methods of explanations are inherently model-specific. This means that the explainer depends on the model architecture and cannot be re-used for other classifier architectures without designing the explanation algorithm specifically for the new architecture as illustrated in Figure \ref{fig:intrinsicmodel}.

\subsubsection{Trees and Rule-based Models}
Shallow rule-based models such as decision trees and decision lists are inherently interpretable. Many explainable algorithms including LIME and SHAP uses linear or tree based models for their globally explainable extensions of the core algorithms. Letham et al. \cite{Letham2015} introduced Bayesian Rule Lists (BRL) which is a generative model that yields a posterior distribution over possible decision lists to improve interpretability while keeping accuracy. 

The rule list has an \textit{if}, \textit{else}, and \textit{elseif} rules generalized as the IF-THEN rule antecedent and predictions. As we add more IF-THEN rules to the decision list, the model becomes more accurate and interpretable. However, support for explanations deteriorate with large number of conditions. One way to simplify the problem is to find the frequent rule patterns and learn a decision list from the distribution using Bayesian techniques.

By picking a sample rule list from the priori distribution and iteratively adding and editing the rules, BRL tries to optimize the rules such that the new rule distribution follows the posteriori distribution. Once optimized, new rules can be sampled from the posteriori distribution. Recent research have improved the scalability of BRL \cite{Yang2017Scalable} by improving the theoretical bounds, computational reuse, and highly tuned language libraries. 

\subsubsection{Generalized additive models (GAMs)}
Caruana et al. \cite{Caruana2015} introduced Generalized additive models (GAMs) with pairwise interactions (GA${^2}$Ms) to improve the accuracy while maintaining interpretability of GAMs. However, for certain models, GAMs require often millions of decision trees to provide accurate results using the additive algorithms. Also, depending on the model architecture, over-regularization reduces accuracy of GAM models which are fit using splines. 
Numerous methods have improved GAMs. Perhaps the most important work is the recent Neural Additive Models we discussed in subsection \ref{subsec:nam}. 

\subsubsection{Sparse LDA and Discriminant Analysis}
A Bayesian non-parametric model, Graph-Sparse LDA, was introduced in \cite{Doshi-Velez2015} to find interpretable, predictive topic summaries to textual categories on datasets with hierarchical labeling. Grosenick et al. \cite{Grosenick2008} introduced a method called Sparse Penalized Discriminant Analysis (SPDA) to improve the spatio-temporal interpretability and classification accuracy of learning algorithms on Functional Magnetic Resonance Imaging (FMRI) data. 

As we see in published research, there are several restrictions to use model intrinsic architectures as it requires careful algorithm development and fine-tuning to the problem setting. The difficulty in using concepts from model intrinsic architectures and apply them in existing high-accuracy models to improve interpretability is a disadvantage of model-intrinsic methods. However, as long as a reasonable performance limit is set, model intrinsic architectures for XAI could help accelerate inherently interpretable models for future AI research.

\subsection{Post-Hoc}

\begin{figure}[!b]
\begin{center}
\includegraphics[width=0.95\linewidth]{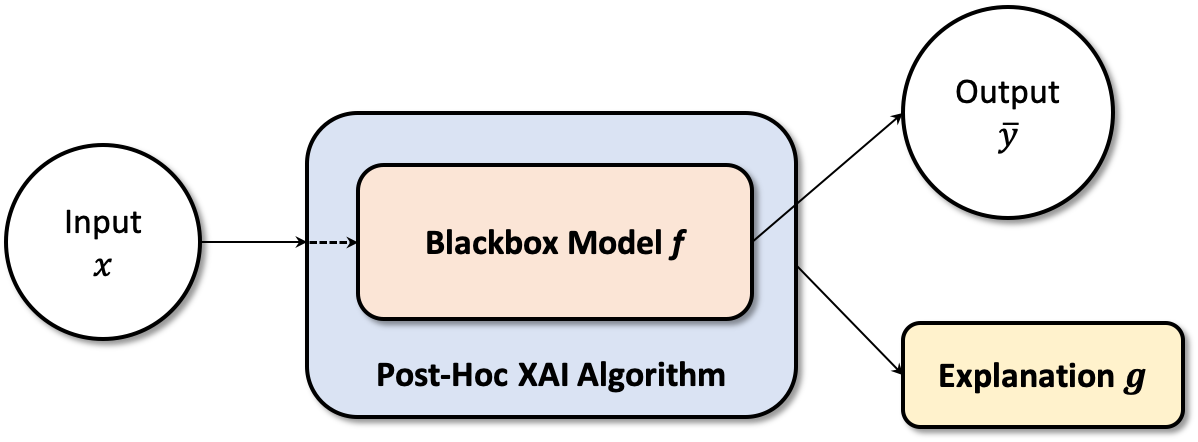}
\end{center}
  \caption{High-level illustration of post-hoc model explainability algorithms. Here, the explainability algorithm is applied on \bm{$f$} such that \bm{$f$} is made explainable externally.}
\label{fig:posthocmodel}
\end{figure}

Explaining pre-trained classifier decisions require algorithms to look at AI models as black or white boxes. A black box means the XAI algorithm doesn't know the internal operations and model architectures. In white box XAI, algorithms have access to the model architecture and layer structures. Post-hoc explanation methodology is extremely useful as existing accurate models can benefit from added interpretability. Most post-hoc XAI algorithms are hence model-agnostic such that the XAI algorithm will work on any network architectures as illustrated in Figure \ref{fig:posthocmodel}. This is one of the main advantages of post-hoc explainable algorithms. For example, an already trained well established neural network decision can be explained without sacrificing the accuracy of the trained model.

Deconvolution network \cite{Zeiler2014} could be used to generate post-hoc explanations of layer-wise activations. Saliency maps \cite{Simonyan2013} and most attribution based methods \cite{Sundararajan2017} are applied considering the network as a white or black box. LRP technique \cite{Bach2015} discussed above is done after training the model completely. Shapley sampling methods \cite{Lundberg2017} are also post-hoc and model agnostic. Activation maximization technique \cite{Erhan2010} is applicable to any network in which we can find gradients values to optimize activations. 

\begin{table*}[!t]
\caption{Summary of published research in explainability and interpretability of deep learning algorithms. \textbf{*} indicates that a preprint version was published an year prior to the conference or journal version.}
\label{tab:papers}
\centering
\begin{tabular}{||l|l|l|l|l|l|p{1.4cm}|l||}
\hline
Method Name & Publication & Year & Scope  & Methodology  & Usage &  Agnostic or Specific & Domain \\ 
\hline\hline
Bayesian averaging over decision trees  & Schetinin et al. \cite{Schetinin2007}              & 2007 & GL     & OT    & IN    & MS    & TAB \\ 
\hline      
SPDA                                    & Grosenick et al. \cite{Grosenick2008}              & 2008 & GL     & OT    & IN    & MS    & TXT \\ 
\hline      
Activation Maximization                 & Erhan et al. \cite{Erhan2010}                  & 2010 & LO     & BP    & PH    & MA    & IMG \\ 
\hline          
Gradient-based Saliency Maps            & Simonyan et al. \cite{Simonyan2013}               & 2013 & LO     & BP    & PH    & MA    & IMG \\ 
\hline          
Bayesian Case Model (BCM)               & Kim et al. \cite{Kim2014}                    & 2014 & GL     & OT    & IN    & MS    & Any \\ 
\hline          
DeConvolutional Nets                    & Zeiler et al. \cite{Zeiler2014}                 & 2013 & LO     & BP    & PH    & MA    & IMG \\ 
\hline          
GAM                                     & Caruana et al. \cite{Caruana2015}                & 2015 & GL     & OT    & IN    & MS    & TAB \\ 
\hline          
LRP                                     & Back et al. \cite{Bach2015}                   & 2015 & Both   & BP    & PH    & MA    & IMG \\ 
\hline
Guided Backprop                         & Springenberg et al. \cite{Springenberg2015}           & 2015 & LO     & BP    & PH    & MA    & IMG \\ 
\hline
Bayes Rule Lists                        & Letham et al. \cite{Letham2015}                 & 2015 & GL     & OT    & IN    & MS    & TAB \\ 
\hline
CAM                                     & Zhou et al. \cite{Zhou2016Learning}           & 2016 & LO     & BP    & PH    & MA    & IMG \\ 
\hline
LIME                                    & Ribeiro et al. \cite{Ribeiro2016}                & 2016 & Both   & PER   & PH    & MA    & Any \\ 
\hline      
Shapley Sampling                        & Lundberg et al. \cite{Lundberg2017}               & 2017 & Both   & PER   & PH    & MA    & Any \\ 
\hline      
Grad-CAM                                & Selvaraju et al. \cite{Selvaraju2017}              & 2017* & LO    & BP    & PH    & MA    & IMG \\ 
\hline      
Prediction Difference Analysis (PDA)    & Zintgraf et al. \cite{Zintgraf2017}               & 2017 & LO     & PER   & PH    & MA    & IMG \\ 
\hline
Deep Taylor Expansion                   & Montavon et al. \cite{Montavon2017}               & 2017 & LO     & OT    & PH    & MA    & IMG \\ 
\hline
Deep Attribution Maps                   & Ancona et al. \cite{Ancona2017}                 & 2017 & LO     & BP    & PH    & MA    & IMG \\ 
\hline
Axiomatic Attributions                  & Sundararajan et al. \cite{Sundararajan2017}           & 2017 & LO     & BP    & PH    & MA    & IMG \\ 
\hline
PatternNet and PatternAttribution       & Kindermans et al. \cite{Kindermans2017Learning}     & 2017 & LO     & BP    & PH    & MA    & IMG \\ 
\hline
Concept Activation Vectors              & Kim et al. \cite{Kim2018Interpretability}    & 2018 & GL     & OT    & PH    & MA    & IMG \\ 
\hline
RISE                                    & Petsiuk et al. \cite{Petsiuk2018}                & 2018 & LO     & PER   & PH    & MA    & IMG \\ 
\hline
Grad-CAM++                              & Chattopadhay et al. \cite{Chattopadhay2018}           & 2018 & LO     & BP    & PH    & MA    & IMG \\ 
\hline
Randomization and Feature Testing       & Burns et al. \cite{Burns2019}                  & 2019 & LO     & PER   & PH    & MA    & IMG \\ 
\hline
Salient Relevance (SR) map              & Li et al. \cite{Li2019}                     & 2019 & LO     & BP    & PH    & MA    & IMG \\ 
\hline
Spectral Relevance Analysis             & Lapuschkin et al. \cite{Lapuschkin2019}             & 2019 & GL     & BP    & PH    & MA    & IMG \\ 
\hline
Global Attribution Mapping              & Ibrahim et al. \cite{Ibrahim2019Global}          & 2019 & GL     & PER   & PH    & MA    & IMG \\ 
\hline
Automatic Concept-based Explanations    & Ghorbani et al. \cite{Ghorbani2019Towards}        & 2019 & GL     & OT    & PH    & MA    & IMG \\ 
\hline
CaCE                                    & Goyal et al. \cite{Goyal2019Explaining}        & 2019 & GL     & OT    & PH    & MA    & IMG \\ 
\hline
Neural Additive Models                  & Agarwal et al. \cite{Agarwal2020Neural}          & 2020 & GL     & OT    & IN    & MS    & IMG \\
\hline

\multicolumn{8}{p{.8\linewidth}}{Global: GL, Local: LO, Others: OT, BackProp: BP, Perturbation: PER, Model-specific: MS, Model-agnostic: MA, Tabular: TAB, Image: IMG, Test: TXT, Any: Image, Text, or Tabular.}

\end{tabular}
\end{table*}

\section{Evaluation Methodologies, Issues, and Future Directions}
\label{sec:evaluation}
So far, we focused on XAI algorithms and methods categorized under scope, methodology, and usage. The seminar works discussed in the survey is tabulated in Table \ref{tab:papers}. A fundamental challenge in XAI research is to evaluate the several proposed algorithms on real-world settings. 
Our survey on evaluation techniques suggested that the field is still immature with primary focus on a human-in-the-loop evaluations. Quantitative general evaluation schemes are yet to be explored. However, we summarize here some of the methods which improve human understandability of explainability method results based on \cite{Doshi2017Towards,Miller2019,Elshawi2019Interpretabillity}. In general, each explanation should follow the below constraints to be usable by humans in a real-world setting:
\begin{enumerate}
    \item Identity or Invariance: Identical data instances must produce identical attributions or explanations.
    \item Stability: Data instances belonging to the same class $c$ must generate comparable explanations $g$.
    \item Consistency: Data instances with change in all but one feature must generate explanations which magnifies the change.
    \item Separability: Data instances from different populations must have dissimilar explanations.
    \item Similarity: Data instances, regardless of class differences, closer to each other, should generate similar explanations.
    \item Implementation Constraints: Time and compute requirement of the explainable algorithm should be minimal.
    \item Bias Detection: Inherent bias in data instances should be detectable from the testing set. Similarity and separability measures help achieve this.
\end{enumerate}

\subsection{Evaluation Schemes}
Several evaluation schemes have been suggested by the research community in the recent years. We present here some of the evaluation techniques that are actively gaining traction from the research community:
\begin{itemize}

\item{System Causability Scale (SCS):}
As the explainability methods are applied to human-facing AI systems which does automated analysis of data, evaluation of human-AI interfaces as a whole is also important. A System Causability Scale (SCS) was introduced in \cite{Holzinger2020} to understand the requirements for explanations of a user-facing human-AI machine-interface, which are often domain specific. Authors described a medical scenario where the SCS tool was applied to Framingham Risk Tool (FRT) to understand the influence and importance of specific characteristics of the human-AI interface.

\item{Benchmarking Attribution Methods (BAM):}
In a preprint publication, \cite{BAM2019} introduced a framework called Benchmarking Attribution Methods (BAM) to evaluate the correctness of feature attributions and their relative importance. A BAM dataset and several models were introduced. Here, the BAM dataset is generated by copying pixel groups, called Common Features (CF), representing object categories from MSCOCO dataset \cite{Lin2014Microsoft} and pasting them to MiniPlaces dataset\cite{Zhou2018Places}. The hypothesis is that, if we have the same pixel group of information in the same spatial location of all of $X$, then the model should ignore it as a feature of relative importance. Hence, attribution methods focusing on pasted objects are simply not doing a good job at enhancing feature attributions of important features. Authors provided model contrast score (MCS) to compare relative feature importance between difference models, input dependence rate (IDR) to learn the dependence of CF on a single instance, and input independence rate (IIR) as a percentage score of images whose average feature attributions $g_r \in \mathbb{R}$ for region $r$ with and without CF is less than a set threshold.

\item{Faithfulness and Monotonicity:}
In \cite{Melis2018Towards}, authors described a metric, named Faithfullness, to evaluate the correlation between importance scores of features to the performance effect of each feature towards a correct prediction. By incrementally removing important features and predicting on the edited data instance, we measures the effect of feature importance and later compare it against the interpreter's own prediction of relevance. 
In \cite{Luss2019Generating}, authors introduce monotonic attribute functions and thus the Monotonicity metric which measures the importance or effect of individual data features on the performance of the model by incrementally adding each feature in the increasing order of importance to find model performance. The model performance is expected to increase as more important features are added.

\item{Human-grounded Evaluation Benchmark:}
In \cite{Mohseni2018Human}, Moshseni et al. introduced a human-grounded evaluation benchmark to evaluate local explanations generated by an XAI algorithm. Authors created a subset of ImageNet dataset \cite{imagenet_cvpr09} and asked human annotators to manually annotate the images for the particular classes. A weighted explanation map was generated which summarized an average human representation of explanations. By comparing the explanations generated by locally explainable algorithms, authors presented a method to understand the precision of XAI explanations compared to human generated explanations. One fundamental flaw of this method could be added human bias in the explanations. However, human labels of individual data points from a large population could nullify the effect of inherent bias.

\end{itemize}

\begin{figure*}[!t]
\begin{center}
\includegraphics[width=\linewidth]{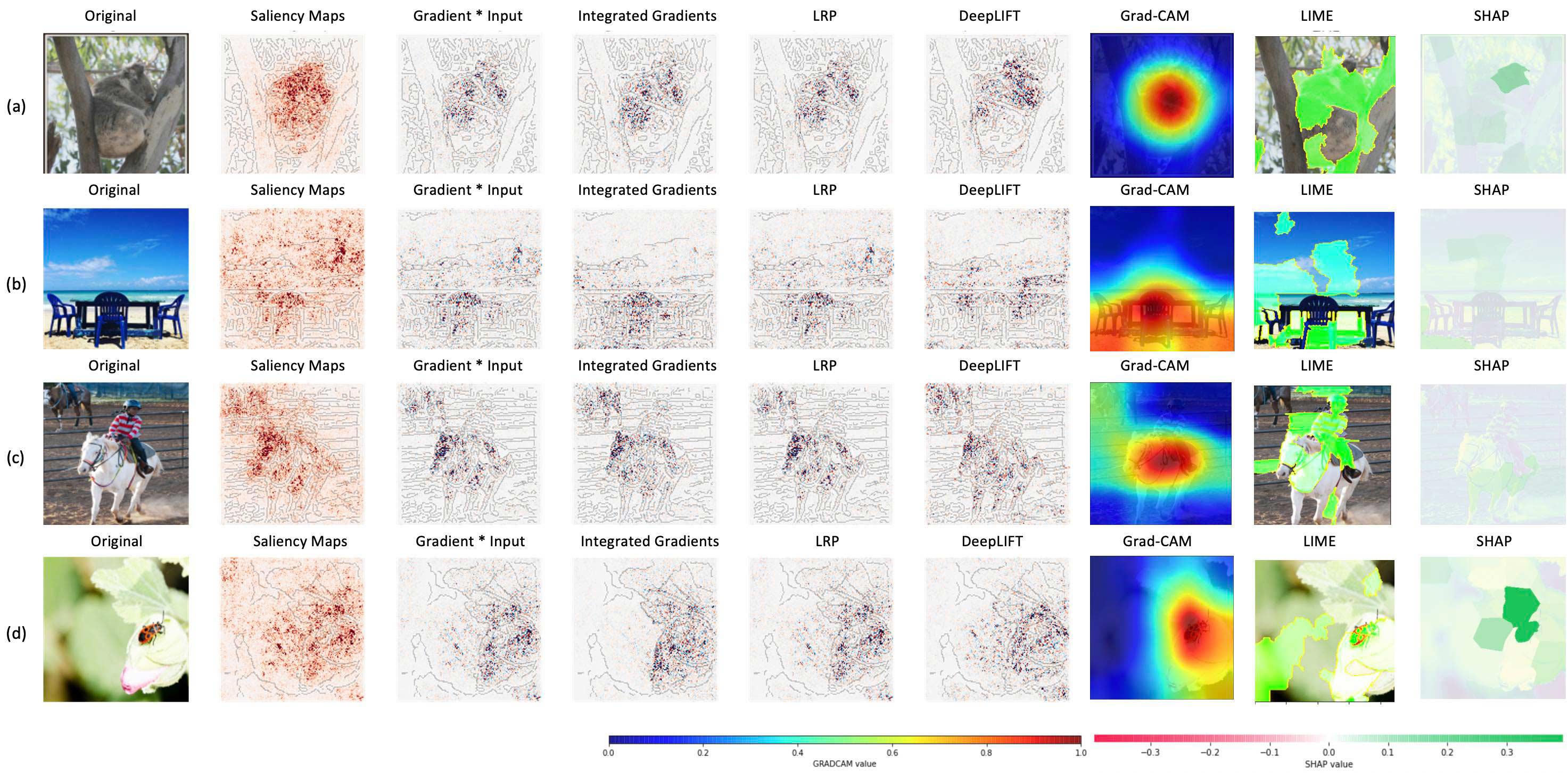}
\end{center}
    \caption{We evaluate different gradient-based and perturbation-based techniques in this figure. LIME and SHAP uses segmented superpixels to understand feature importance, while gradient based saliency maps, Integrated Gradients, LRP, DeepLIFT, and Grad-CAM use backpropagation based feature importance in a pixel level. Original prediction accuracies of a pre-trained InceptionV3 model on the images in each rows provided are as follows: (a) `koala', $94.5\%$, (b) `sandbar', $38.0\%$, (c) `arabian camel', $17.4\%$, and (d) `leaf beetle', $95.5\%$. Each column represents the attribution map generated by individual XAI methods. Scales to assess Grad-CAM and SHAP values are provided in the lower right section of the image. Gradient visualizations of this figure are created using DeepExplain package while visualizations for Grad-CAM, LIME, and SHAP are created by their own individual implementations. The experiments were carried out in Jetstream cloud \cite{Stewart2015Jetstream}. This image is better viewed in color.}
\label{fig:evaluatioxai}
\end{figure*}

\subsection{Software Packages}
OpenSource packages have greatly improved reproducible research and has been a real boon to recent research in deep learning and XAI alike. We mention here some XAI software packages available in GitHub.
\begin{itemize}
    \item \textbf{Interpret} by InterpretML can be used to explain blackbox models and currently supports explainable boosting, decision trees, decision rule list, linear\/logistic regression, SHAP kernel explainer, SHAP tree explainer, LIME, morris sensitivity analysis, and partial dependence. Available at \url{https://github.com/interpretml/interpret}.
    \item \textbf{IML} package \cite{Molnar2018Iml} is maintained by Christoph Molnar, author of \cite{Molnar2019}. The package covers feature importance, partial dependence plots, individual conditional expectation plots, accumulated local effects, tree surrogates, LIME, and SHAP. Available at \url{https://github.com/christophM/iml}.
    \item \textbf{DeepExplain} package is maintained by Marco Ancona, author of \cite{Ancona2017}. The package supports various gradient-based techniques such as saliency maps, gradient\*input, integrated gradients, DeepLIFT, LRP, etc. and perturbation-based methods such as occlusion, SHAP, etc. Available at \url{https://github.com/marcoancona/DeepExplain}.
    \item \textbf{DrWhy} by ModelOriented is a package with several model agnostic and model specific XAI techniques including feature importance, ceteris paribus, partial dependency plots, conditional dependency, etc. Available at \url{https://github.com/ModelOriented/DrWhy}
\end{itemize}

\subsection{A Case-study on Understanding Explanation Maps}
In Figure \ref{fig:evaluatioxai}, we illustrate the explanation maps generated using various gradient- and perturbation-based XAI techniques for four images from ImageNet \cite{imagenet_cvpr09} dataset to explain the decisions an InceptionV3 model pre-trained on ImageNet. Here, each row starts with an original image from ImageNet followed by explanation map generated by gradient algorithms such as 1) saliency maps, 2) gradient times input, 3) integrated gradients, 4) LRP, 5) DeepLIFT, and 6) GradCAM, and perturbation-based techniques such as 1) LIME and 2) SHAP. 

GradCAM generates a heatmap of values ranging from 0 to 1, where 0 means no influence and 1 means highest influence of individual pixels towards the model output decision. Similarly, SHAP method follows a scale for SHAP values. However, SHAP scale ranges from -0.3 to +0.3 indicating that negative values decrease output class probability and positive values increase the output class probability for the corresponding input. Here 0.3 is the largest SHAP value generated for the set of four images considered. Gradient visualizations of this figure are created using DeepExplain package while visualizations for Grad-CAM, LIME, and SHAP are created by their own individual implementations.

Original image column of row (a) in Figure \ref{fig:evaluatioxai} indicates a correct prediction of an image of a Koala with $94.5\%$ prediction accuracy, row (b) indicates a correct prediction of a sandbar image with $38.0\%$ accuracy, row (c) indicates an incorrect prediction of a horse as an arabian camel with $17.4\%$ accuracy, and row (d) indicates correct prediction of a leaf beetle with $95.5\%$ percentage accuracy. We then compare the explanation maps, in different columns, generated by various XAI techniques as discussed above.

\begin{figure*}[!t]
\begin{center}
\includegraphics[width=0.9\linewidth]{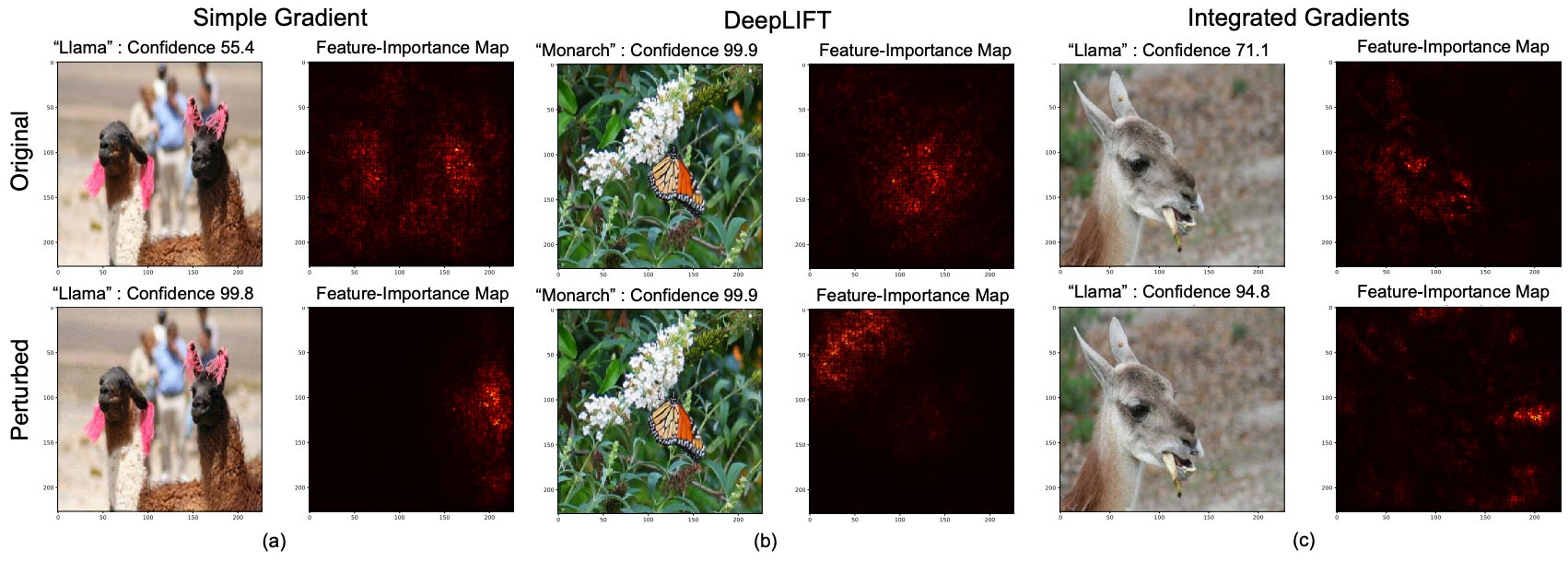}
\end{center}
  \caption{Illustration from \cite{Ghorbani2019Interpretation} showing adversarial attacks involving small perturbations to input layer of neural network. We see that small perturbations doesn't affect the accuracy of predictions. However, feature importance maps are highly affected by the small changes. This illustrates the flaws in current gradient-based techniques.}
\label{fig:advfeatimp}
\end{figure*}

Focusing on saliency maps, gradient times input, and integrated gradients in Figure \ref{fig:evaluatioxai}, we can visually verify the improvements achieved by integrated gradients over the prior gradient-based methods. This is apparent in the images with lower class probabilities. For example, in row (b), we can verify that the integrated gradients generated high attributions around the sandy beach, plastic chairs, and a little bit of the blue sky. As human evaluators, we can make sense of this output because human experience suggests that a sandbar involve a beach, hopefully on a sunny day with bright blue clouds. A stark difference is apparent in Grad-CAM visualizations where the class output generated a heatmap which is focused primarily on the plastic chair and sandy beach, without much emphasis on the clouds. Perturbation-based methods such as LIME and SHAP generated superpixels which maximized the class probability. Here, we see that LIME is focusing on primarily the chairs and the sky, whereas SHAP is focusing on the beach and the sky. We also note that SHAP values generated are very low, indicating lesser influence to the confidence score. 

\subsection{Limitations of XAI Visualizations and Future Directions}

The discussion above brings some important flaws of XAI visualizations and interpretability techniques - 1) the inability of human-attention to deduce XAI explanation maps for decision-making, and 2) unavailability of a quantitative measure of completeness and correctness of the explanation map. This suggests that the further use of visualization techniques for mission-critical applications must be reconsidered moving forward. Also, better ways of representing and presenting explanations should be considered. For example, in \cite{Weerts2019Human}, Weerts et al. studied the impact of SHAP explanations in improving human performance for alert processing tasks. The authors presented a human-grounded study to evaluate whether certain decision-making scenarios can be improved by providing explanations to decisions. Results showed that additional SHAP explanations to class output probability did not improve the decision-making of individuals. Authors saw more interest in final class score for making decisions which could be catastrophic in mission-critical scenarios.

Similarly, in \cite{Mohseni2018Human}, Mohseni et al. presented a human-grounded evaluation benchmark and evaluated performance of LIME algorithm by comparing the explanation map generated by LIME to that of weighted explanation map of 10 human annotations. Results suggested that LIME creates some attributions irrelevant to human explanations which causes low explanation precision compared to weighted explanation map generated by human annotators. This sheds light to the importance of understanding the mode of explanations as application-grounded, human-grounded, and functionally-grounded explanations \cite{Doshi2017Towards} to improve explanation maps by meta information generated by humans, adding more constraints to explanations, or introducing formal definitions of explanations to the optimization problem.

Several other flaws of explanation map visualization are explained by researchers in recent publications. In \cite{Ghorbani2019Interpretation}, Ghorbani et al. showed that small perturbations on the input instance generate large changes in the output interpretations that popular XAI methods generate. These \textit{adversarial} examples, thus threw off the interpretable saliency maps generated by popular methods such as DeepLIFT and Integrated Gradients. This is illustrated in Figure \ref{fig:advfeatimp}. Additionally, in \cite{Wang2019Bias}, Wang et al. showed that bias term which is often ignored could have high correlations towards attributions.

In \cite{Kindermans2019Reliability}, Kindermans et al. explained that explanations of networks are easily manipulable by simple transformations. Authors note that expressiveness of Integrated Gradients \cite{Sundararajan2017} and Deep Taylor Decomposition \cite{Montavon2017} highly depend on the reference point, for example a baseline image \bm{$x^{'}$}, and suggest that the reference point should be a hyperparameter instead of being determined \textit{a priori}. Authors mentioned that most gradient-based methods attribute incorrectly to constant vector transformations and that input invariances should be a prerequisite for reliable attributions.

In \cite{Adebayo2018Sanity}, Adebayo et al. suggested that gradient-based methods are inherently dependent on the model and data generating process. Authors proposed two randomization tests for gradient methods namely model parameter randomization test and data randomization test. Model parameter randomization test compared the output of saliency method for a trained model versus the same model with random weights. Data randomization test applied the same saliency method for an input instance and the same instance with a set of invariances. Authors found that Gradients and GradCAM passed the sanity checks while Guided Backprop and Guided GradCAM methods failed the tests suggesting that these methods will generate some explanations even without proper training.

Newer methods proposed in literature such as explaining with Concepts \cite{Kim2018Interpretability, Ghorbani2019Towards, Goyal2019Explaining} which we discussed in subsection \ref{subsec:tcav} could be viewed as a new class of meta-explanations which improve both perturbation- and gradient-based XAI methods. By exploring explanations as concepts, one could have additional meta information on the factors which contributed to individual class predictions along with traditional explanation by locally explainable algorithms. 

In \cite{Zhou2018Interpretable}, Zhou et al. introduced Interpretable Basis Decomposition as a way of decomposing individual explanation based on different objects or scenes in the input instance. By decomposing the decision to several individual concept explanations, IBD could help evaluate importance of each concepts towards a particular decision.

In \cite{Kindermans2017Learning}, Kindermans et al. suggested improvements to gradient-based methods and proposed PatternNet and PatternAttribution which can estimate the component of the data that caused network activations. Here, PatternNet is similar to finding gradients but is instead done using a layer-wise backprojection of the estimated signal (data feature) to the input space. PatternAttribution improves upon LRP to provide a neuron-wise attribution of input signal to the corresponding output class.

\section{Conclusion}
\label{sec:conclusion}
Blindly trusting the results of a highly predictive classifier is, by today's standard, inadvisable, due to the strong influence of data bias, trustability, and adversarial examples in machine learning. In this survey, we explored why XAI is important, several facets of XAI, and categorized them in respect of their scope, methodology, usage, and nature towards explaining deep neural network algorithms. A summary of the seminal algorithms explained in the survey are tabulated in Table \ref{tab:papers}.

Our findings showed that considerable research in XAI is focused in model-agnostic post-hoc explainability algorithms due to their easier integration and wider reach. Additionally, there is a large interest in additive and local surrogate models using superpixels of information to evaluate the input feature attributions. Researchers are uncovering limitations of explanation maps visualizations and we see a shift from local perturbation- and gradient-based models due to their shortcomings in adversarial attacks and input invariances. A new trend in using concepts as explanations are gaining traction. However, evaluating these methods are still a challenge and pose an open question in XAI research.

Current research landscape in XAI evaluation illustrates that the field of XAI is still evolving and that XAI methods should be developed and chosen with care. User studies have shown that typical explanation maps alone might not aid in decision making. Human bias in interpreting visual explanations could hinder proper use of XAI in mission-critical applications. Recent developments in human-grounded evaluations shows promising improvements to the XAI evaluation landscape.

\bibliographystyle{IEEEtran}
\bibliography{references}

\begin{IEEEbiography}
[{\includegraphics[width=1in,height=1.25in,clip,keepaspectratio]{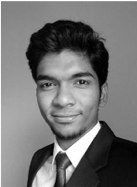}}]{Arun Das} is currently a Ph.D. student and research fellow at the Secure AI and Autonomy Lab and Open Cloud Institute of University of Texas at San Antonio (UTSA), San Antonio, TX, USA. Arun received the Bachelor of Technology (B.Tech.) degree in Electrical and Electronics Engineering from Cochin University of Science and Technology, Kerala, India, in 2013 and the M.S. degree in Computer Engineering from the University of Texas at San Antonio, San Antonio, TX, USA in 2016. He was a recipient of the UTSA Brain Health Consortium Graduate Student Seed Grant in 2020 for his work in behavior analytics for children with neurotypical disabilities. He is a member of the IEEE, and IEEE Eta Kappa Nu honor society. Arun's research interests are in the areas of artificial intelligence, computer vision, distributed and parallel computing, cloud computing, and computer architecture.
\end{IEEEbiography} 

\begin{IEEEbiography}[{\includegraphics[width=1in,height=1.25in,clip,keepaspectratio]{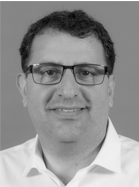}}]{Peyman Najafirad (Paul Rad)} received the PhD. degree in electrical and computer engineering on cyber analytics from the University of Texas at San Antonio, San Antonio, TX, USA. He is the founder and director of the Secure AI and Autonomy Lab, and an Associate Professor with the information systems and cyber security (ISCS) from University of Texas at San Antonio. He has received his first B.S. degree from Sharif University of Technology in Computer Engineering in 1994, his 1st master in artificial intelligence from the Tehran Polytechnic, the 2nd master in computer science from the University of Texas at San Antonio (Magna Cum Laude) in December 1999, and the Ph.D. in electrical and computer engineering from the University of Texas at San Antonio. He was a recipient of the Most Outstanding Graduate Student in the College of Engineering, 2016, Achieving Rackspace Innovation Mentor Program Award for establishing Rackspace patent community board structure and mentoring employees, 2012, Achieving Dell Corporation Company Excellence (ACE) Award in Austin for exceptional performance and innovative product research and development contributions, 2007, and Dell Inventor Milestone Award, Top 3 Dell Inventor of the year, 2005.He holds 15 U.S. patents on cyber infrastructure, cloud computing, and big data analytics with over 300 product citations by top fortune 500 leading technology companies such as Amazon, Microsoft, IBM, Cisco, Amazon Technologies, HP, and VMware. He has advised over 200 companies on cloud computing and data analytics with over 50 keynote presentations. He serves on the advisory board for several startups, high performance cloud group chair at the Cloud Advisory Council (CAC), OpenStack Foundation Member, the number 1 open source cloud software, San Antonio Tech Bloc Founding Member, Children’s Hospital of San Antonio Foundation board member.
\end{IEEEbiography}

\vfill

\end{document}